\pdfoutput=1
\documentclass{article}
\usepackage{nips13submit_e,times}
\usepackage{amsmath,amsfonts}
\usepackage{graphicx}
\usepackage{color}
\usepackage{threeparttable}
\usepackage{multirow}
\usepackage{makecell}
\usepackage{enumitem}
\usepackage{tcolorbox}
\usepackage{colortbl}
\usepackage{mathrsfs}
\usepackage{url}
\usepackage{booktabs}
\usepackage{longtable}
\definecolor{tabcolor}{RGB}{222,234,246}
\definecolor{mcolor}{RGB}{204,255,153}
\definecolor{scolor}{RGB}{255,255,102}
\definecolor{fcolor}{RGB}{255,204,102}

\title{Neural Machine Reading Comprehension:\\ Methods and Trends}
\author{
Shanshan Liu\footnotemark[2], Xin Zhang\footnotemark[3], Sheng Zhang\footnotemark[2], Hui Wang\footnotemark[2], Weiming Zhang\footnotemark[2]\\
Science and Technology on Information Systems Engineering Laboratory\\
College of Systems Engineering\\
National University of Defense Technology\\
\texttt{†\{liushanshan17,zhangsheng,huiwang,wmzhang\}@nudt.edu.cn} \\
\texttt{‡Correspondence:ijunzhanggm@gmail.com}}

\nipsfinalcopy

\begin{document}
\maketitle
\begin{abstract}

Machine reading comprehension (MRC), which requires a machine to answer questions based on a given context, has attracted increasing attention with the incorporation of various deep-learning techniques over the past few years. Although research on MRC based on deep learning is flourishing, there remains a lack of a comprehensive survey summarizing existing approaches and recent trends, which motivated the work presented in this article. Specifically, we give a thorough review of this research field, covering different aspects including (1) typical MRC tasks: their definitions, differences, and representative datasets; (2) the general architecture of neural MRC: the main modules and prevalent approaches to each; and (3) new trends: some emerging areas in neural MRC as well as the corresponding challenges. Finally, considering what has been achieved so far, the survey also envisages what the future may hold by discussing the open issues left to be addressed.
\end{abstract}

\section{Introduction}

\begin{figure}[h]
	\centering
	\includegraphics[width=6.5cm]{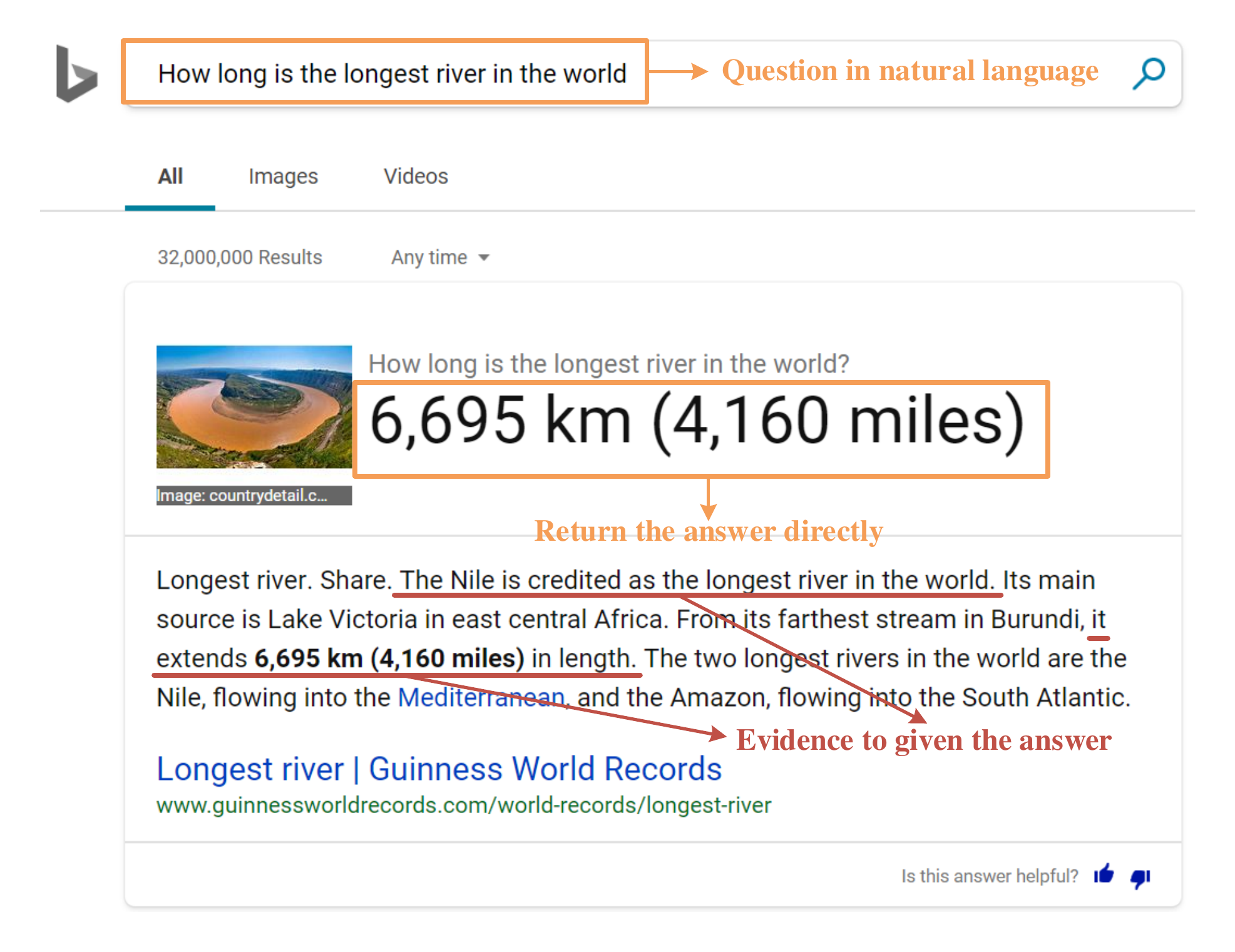}
	\caption{An example of search engine \textit{Bing} (\url {https://cn.bing.com/}) with machine reading comprehension (MRC) techniques.} 
	\label{search}
\end{figure}

Machine reading comprehension (MRC) is a task introduced to test the degree to which a machine can understand natural languages by asking the machine to answer questions based on a given context, which has the potential to revolutionize the way in which humans and machines interact with each other. For example, as shown in Figure \ref{search}, a search engine with MRC techniques can directly return the~correct answers to questions posed by users in natural language rather than a series of related web pages. In addition, smart assistants equipped with an MRC system can read help documents and provide users with high-quality consulting services. In sum, MRC is a promising task, which can make information retrieval more efficient. 

\par Early MRC systems date back to the 1970s, the most notable of which was the QUALM system proposed by Lehnert \cite{lehnert1977process}. However, restricted by its small scale and specific domain, this system could not be widely applied. Research on MRC was ignored in the 1980s and 1990s. In 1999, \mbox{Hirschman et al.} \cite{hirschman1999deep} released an MRC dataset, containing stories of third- to sixth-grade material with five ``wh'' (what, where, when, why, and who) questions, which again directed the research attention to MRC tasks. At that time, methods for solving MRC tasks were mainly rule-based or machine-learning-based: Hirschman et al. \cite{hirschman1999deep} proposed a bag-of-words technique that represented sentences with questions and context as sets of words and chose words occurring in both the question and context as the answer. Riloff et al. \cite{riloff2000rule} designed a rule-based MRC system, Quarc, 
containing different heuristic rules with morphological analysis, such as part-of-speech tagging, semantic class tagging, and entity recognition, for different types of ``wh'' questions. Poon et al. \cite{poon2010machine} used a machine-learning method combining bootstrapping, Markov logic, and self-supervised learning for machine reading. However, those methods suffer from some, if not all, of the following limitations. First, they are mainly based on hand-crafted rules or features that require substantial human effort. Second, those systems are incapable of generalization, and their performance may degrade due to large-scale datasets of myriad types of articles. Finally, some traditional approaches not only ignore long-range dependencies but also fail to extract contextual information.

\begin{figure}[h]
	\centering
	\includegraphics[width=7cm]{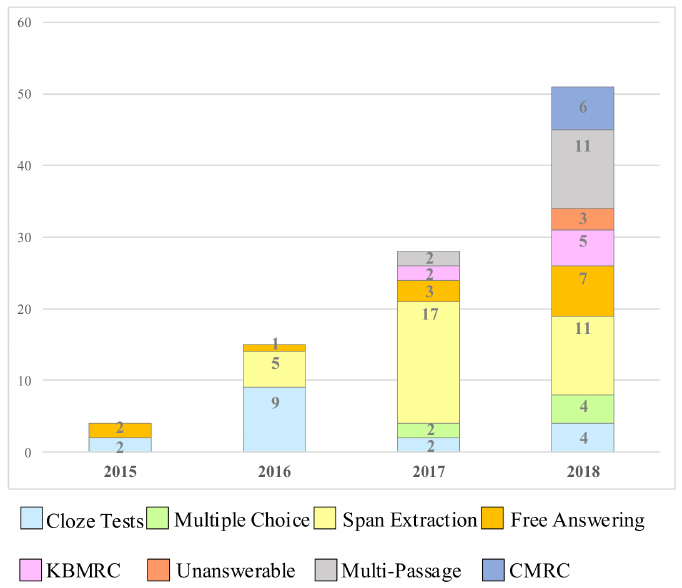}
	\caption{The number of research articles concerned with neural MRC in this survey. KBMRC: knowledge-based machine reading comprehension; CMRC: conversational machine reading comprehension.}
	\label{year}
\end{figure}

Due to the small size of human-generated datasets and the limitations of rule-based and machine-learning-based methods, early MRC systems did not perform well and hence could not be used in practical applications. This situation has changed since 2015, which can be attributed to two driving forces. On the one hand, MRC based on deep learning, also called neural machine reading comprehension, shows its superiority in capturing contextual information and dramatically outperforms traditional systems. On the other hand, a variety of large-scale benchmark datasets, such as CNN \& Daily Mail \cite{hermann2015teaching}, Stanford Question-Answering Dataset (SQuAD) \cite{rajpurkar2016squad} and MS MARCO \cite{nguyen2016ms}, make it possible to solve MRC tasks with deep neural architectures and provide testbeds to extensively evaluate the performance of MRC systems. To illustrate the development trends of neural MRC more clearly, we conducted a statistical analysis of representative articles in this field, and the result is presented in Figure \ref{year}. \mbox{As~shown~in} this figure, on the whole, the number of articles increased exponentially from 2015 to the end of 2018. Furthermore, as time goes on, the types of MRC tasks are becoming increasingly diverse. All these demonstrate that neural MRC is under rapid development and has become the research focus of both academia (Stanford, Carnegie Mellon, etc.) and industry (Google, Facebook, Microsoft, etc.) demonstrated by a large number of authors coming from these institutions.

\begin{figure}[h]
	\centering
	\includegraphics[width=12cm]{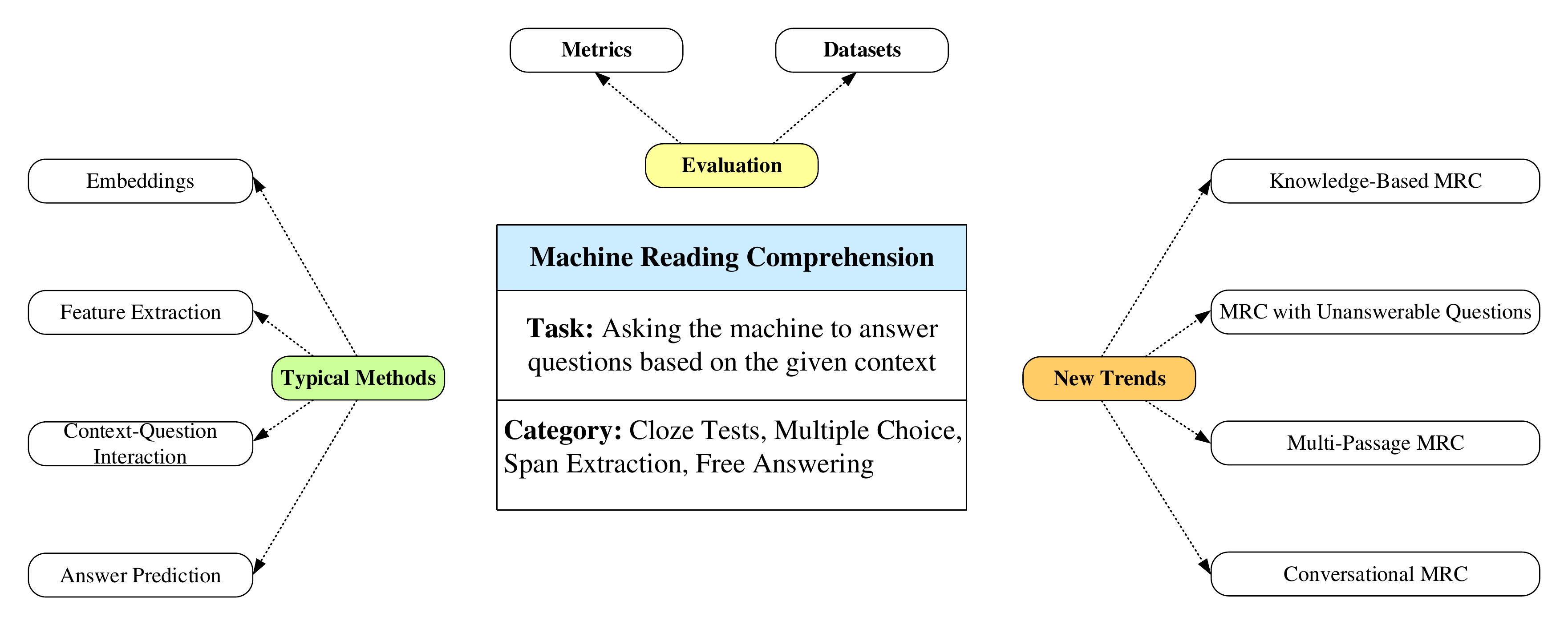}
	\caption{Facets of machine reading comprehension reflected in the structure of this article.}
	\label{summary}
\end{figure}

The flourishing research of neural MRC calls for surveys to systemically study and analyze the recent successes. However, such a comprehensive review still cannot be found. Though Qiu et al. \cite{qiu2019survey} recently give a brief overview to illustrate how to use deep-learning methods to deal with MRC tasks by introducing several classic neural MRC models, they neither give specific definitions of different MRC tasks nor compare them in depth. Moreover, they do not discuss the new trends and open issues in this field. Motivated by the lack of published surveys, we conducted a thorough literature review of recent progress in neural MRC with the expectation that it would help researchers, \mbox{in~particular} newcomers, to obtain a panoramic view of this field. To achieve that goal, we collected papers mainly using Google Scholar (\url {http://scholar.google.com}) with keywords including machine reading comprehension, machine comprehension, reading comprehension, deep learning, and neural networks. \mbox{We selected} from the search results only papers published on related high-profile conferences such as ACL, EMNLP, NAACL, ICLR, AAAI, IJCAI and CoNLL, and restricted the time range to 2015--2018. \mbox{In~addition}, arXiv (\url {http://arxiv.org/}), which includes some of the latest pre-print articles, was used as a supplementary source. Based on more than 85 frequently cited papers collected, a possible organization of the various facets of MRC as a field is shown in Figure \ref{summary}. Our article follows a similar structure. First, we group common MRC tasks into four types according to Chen's categorization in her PhD thesis \cite{chen2018neural}: \textit{cloze tests}, \textit{multiple choice}, \textit{span extraction}, and \textit{free answering}. We further extend this taxonomy by giving a formal definition to each type,  comparing these tasks in different dimensions (Section \ref{s2}). Second, we present the general architecture of neural MRC systems, which consists of four modules: \textit{embeddings}, \textit{feature extraction}, \textit{context-question interaction} and \textit{answer prediction}. \mbox{The~prevalent deep} learning techniques used in each module are also detailed (Section \ref{s3}). Third, some representative datasets and evaluation metrics are described according to different tasks (Section \ref{s4}). Fourth, some new trends, such as \textit{knowledge-based MRC}, \textit{MRC with unanswerable questions}, \textit{multi-passage MRC} and \textit{conversational MRC}, are revealed by figuring out their challenges and describing existing approaches and limitations (Section \ref{s5}). Finally, several open issues are discussed with the hope of shedding light on possible future research directions (Section \ref{s6}). Before proceeding on the next section, as there are lots of terms mentioned in this article, to improve the readability for newcomers, definitions of terms can be found in a glossary presented in Appendix \ref{appendix}.

\section{Tasks} \label{s2}
Machine reading comprehension (MRC) is a basic task of textual question answering (QA), \mbox{in~which} each question is given related context from which  to infer the answer. The objective of MRC is to extract the correct answer from the given context or even generate a more complex answer based on the context. MRC holds the promise of bridging the gap of understanding natural language between humans and machines. The formal definition of MRC is as shown in Table \ref{d_mrc}:

\begin{table}[h]
	\renewcommand{\arraystretch}{1.5}
	\centering
	\caption{Definition of machine reading comprehension.}
	\label{d_mrc}
	\begin{tabular}{p{12cm}}
		\noalign{\hrule height 1pt}
		\bfseries{Machine Reading Comprehension}\\
		\hline
		\rowcolor{tabcolor}Given the context $C$ and question $Q$, machine reading comprehension tasks ask the model to give the correct answer $A$ to the question $Q$ by learning the function $\mathcal{F}$ such that $A=\mathcal{F}(C,Q)$.\\
		\noalign{\hrule height 1pt}
	\end{tabular}
\end{table}

\vspace{12pt}

In this section, following Chen \cite{chen2018neural}, we categorize MRC into four tasks, mainly based on the answer form: cloze tests, multiple choice, span extraction and free answering. For better understanding, some examples of representative datasets are presented in Table \ref{dataset}. In the following part, we give a description of each task with a formal definition and compare them from different dimensions.

\subsection{Cloze Tests}

\par Cloze tests, also known as gap-filling tests, are commonly adopted in exams to evaluate students' language proficiency. Inspired by that, this task is used to measure the ability of machines to understand natural language. In cloze tests, questions are generated by removing some words or entities from a passage. To answer questions, one is asked to fill in the blank with the missing items. Some tasks provide candidate answers, but this is optional. Cloze tests, which add obstacles to reading, require understanding of context as well as usage of vocabulary and are challenging for machine reading comprehension. 
\par The features of cloze tests are as follows: (i) answer $A$ is a word or entity in the given context $C$; and (ii) question $Q$ is generated by removing a word or entity from the given context $C$ such that $Q=C-{A}$. According to such features, cloze tests can be formulated as shown in Table \ref{d_cloze}:

\begin{table}[!h]
	\renewcommand{\arraystretch}{1.5}
	\centering
	\caption{Definition of cloze tests.}
	\label{d_cloze}
	\begin{tabular}{p{12cm}}
		\noalign{\hrule height 1pt}
		\bfseries{Cloze Tests}\\
		\hline
		\rowcolor{tabcolor}Given the context $C$, from which a word or an entity $A (A\in C)$ is removed, the cloze tests ask the model to fill in the blank with the right word or entity $A$ by learning the function $\mathcal{F}$ such that $A=\mathcal{F}(C-\{A\})$.\\
		\noalign{\hrule height 1pt}
	\end{tabular}
\end{table}

\subsection{Multiple Choice}
\par  Multiple choice is another machine reading comprehension task inspired by language proficiency exams. It requires selecting the correct answer to the question from candidates according to the provided context. Compared to cloze tests, answers for multiple choice are not limited to words or entities in the context, so the answer form is more flexible, but it is required for this task to provide candidate answers.
\par The feature of multiple choice is that a list of candidate answers $A=\{A_{1},A_{2},\cdots,A_{n}\}$ is given that can act as auxiliary information to predict the answer. It can be formulated as shown in Table \ref{d_multiple}:
\vspace{12pt}

\begin{table}[h]
	\renewcommand{\arraystretch}{1.5}
	\centering
	\caption{Definition of multiple choice.}
	\label{d_multiple}
	\begin{tabular}{p{12cm}}
		\noalign{\hrule height 1pt}
		\bfseries{Multiple Choice}\\
		\hline
		\rowcolor{tabcolor}Given the context $C$, the question $Q$ and a list of candidate answers $A=\{A_{1},A_{2},\cdots,A_{n}\}$, the multiple-choice task is to select the correct answer $A_{i}$ from $A$ $(A_{i} \in A)$ by learning the function $\mathcal{F}$ such that $A_{i}=\mathcal{F}(C,Q,A)$\\
		\noalign{\hrule height 1pt}
	\end{tabular}
\end{table}


\subsection{Span Extraction}
Although cloze tests and multiple choice can measure the machine's ability to understand natural language to some extent, there are limitations in those tasks. More specifically, words or entities are not sufficient to answer questions. Instead, some complete sentences are required. Moreover, there are no candidate answers in many cases. The span extraction task can  overcome these weaknesses. Given the context and the question, this task asks the machine to extract a span of text from the corresponding context as the answer. 

\par The feature of span extraction is that answer $A$ is required to be a continuous subsequence \mbox{of the given} context $C$. It can be formulated as shown in Table \ref{d_span}:
\vspace{12pt}

\begin{table}[h]
	\renewcommand{\arraystretch}{1.5}
	\centering
	\caption{Definition of span extraction.}
	\label{d_span}
	\begin{tabular}{p{12cm}}
		\noalign{\hrule height 1pt}
		\bfseries{Span Extraction}\\
		\hline
		\rowcolor{tabcolor}Given the context $C$, which consists of $n$ tokens, that is $C=\{t_{1},t_{2},\cdots,t_{n}\}$, and the question $Q$, the span extraction task requires extracting the continuous subsequence $A={\{t_{i},t_{i+1},\cdots,t_{i+k}\}} (1 \leq i \leq i+k \leq n)$ from context $C$ as the correct answer to question $Q$ by learning the function $\mathcal{F}$ such that $A=\mathcal{F}(C,Q)$.\\
		\noalign{\hrule height 1pt}
	\end{tabular}
\end{table}


\subsection{Free Answering}
Compared to cloze tests and multiple choice, the span extraction task makes great strides \mbox{in allowing} machines to give more flexible answers, yet it is not enough, because giving answers restricted to a span of the context is still unrealistic. To answer the questions, the machine needs to reason across multiple pieces of the text and summarize the evidence. Among the four tasks, free answering is the most complicated, as there are no limitations to its answer forms, and it is more suitable for real application scenarios.
\par In contrast to the other tasks, free answering reduces some constraints and focuses much more on using free-form natural language to better answer questions and it can be formulated as shown in Table \ref{d_free}:
\vspace{12pt}

\begin{table}[h]
	\renewcommand{\arraystretch}{1.5}
	\centering
	\caption{Definition of free answering.}
	\label{d_free}
	\begin{tabular}{p{12cm}}
		\noalign{\hrule height 1pt}
		\bfseries{Free Answering}\\
		\hline
		\rowcolor{tabcolor}Given the context $C$ and the question $Q$, the correct answer $A$ in free answering task may not be subsequence in the original context $C$, namely either $A \subseteq C$ or $A \not\subseteq C$. The task requires predicting the correct answer $A$ by learning the function $\mathcal{F}$ such that $A=\mathcal{F}(C,Q)$.\\
		\noalign{\hrule height 1pt}
	\end{tabular}
\end{table}

\vspace{50pt}

\begin{longtable}[c]{cll}
	\caption{\upshape A few examples of MRC datasets}
	\label{dataset}\\
	\hline
	\rowcolor{tabcolor}\multicolumn{3}{l}{\textbf{Cloze Tests}}\\ 
	\hline
	\multirow{4}{*}{\makecell[c]{CNN \& \\ Daily Mail \cite{hermann2015teaching}}} & Context: & \makecell[l]{the \textit{ent381} producer allegedly struck by \textit{ent212}\\ will not  press charges against the ``\textit{ent153}'' host,\\ his lawyer said Friday. \textit{ent212}, who hosted one\\ of the most-watched television shows in the world,\\ was dropped by the \textit{ent381} Wednesday after an\\ internal investigation by the \textit{ent180} broadcaster\\ found he had subjected producer \textit{ent193} ``to an\\ unprovoked physical and verbal attack.''}\\
	\cmidrule{2-3}
	&Question: & \makecell[l]{producer \underline{\ \ \textbf{X}\ \ } will not press charges against\\ \textit{ent212}, his lawyer says.}\\ 
	\cmidrule{2-3}
	&Answer: & \textit{ent193}\\
	\hline
	\rowcolor{mcolor}\multicolumn{3}{l}{\textbf{Multiple Choice}}\\
	\hline
	\multirow{3}{*}{RACE \cite{lai2017race}}&Context: & \makecell[l]{If you have a cold or flu, you must always deal\\ with used tissues carefully. Do not leave dirty\\ tissues on your desk or on the floor. Someone\\ else must pick these up and viruses could be\\ passed on.}\\
	\cmidrule{2-3}
	&Question:& \makecell[l]{Dealing with used tissues properly is important\\ because \_\_\_\_\_.}\\
	\cmidrule{2-3}
	&Options:& \makecell[l]{A. it helps keep your classroom tidy\\B. people hate picking up dirty tissues\\ \color{blue}C. it prevents the spread of colds and flu\\ D. picking up lots of tissues is hard work}\\
	\cmidrule{2-3}
	&Answer:& C\\
	\hline
	\rowcolor{scolor}\multicolumn{3}{l}{\textbf{Span Extraction}}\\
	\hline
	\multirow{3}{*}{SQuAD \cite{rajpurkar2016squad}} & Context: &\makecell[l]{Computational complexity theory is a branch of\\ the theory of computation in theoretical computer\\ science that focuses on classifying computational\\ problems according to their \color{blue}inherent difficulty\color{black},\\ and relating those classes to each other. A compu-\\tational problem is understood to be a task that is\\ in principle amenable to being solved by a computer,\\ which is equivalent to stating that the problem may\\ be solved by mechanical application of mathematical\\ steps, such as an algorithm. } \\
	\cmidrule{2-3}
	&Question: & \makecell[l]{By what main attribute are computational problems\\ classified using computational complexity theory?} \\ 
	\cmidrule{2-3}
	&Answer:&inherent difficulty \\
	\hline
	\rowcolor{fcolor}\multicolumn{3}{l}{\textbf{Free Answering}}\\
	\hline
	\multirow{7}{*}{MS MARCO \cite{nguyen2016ms}}& Context 1:& \makecell[l]{Rachel Carson's essay on The Obligation to Endure,\\ is a very convincing argument about the harmful\\ uses of chemical, pesticides, herbicides and fertilizers\\ on the environment.} \\
	& $\cdots\cdots$ & \\
	&Context 5: & \makecell[l]{Carson believes that as man tries to eliminate\\ unwanted insects and weeds; however he is actually\\ causing more problems by polluting the environment\\ with, for example, DDT and harming living things}.\\
	& $\cdots\cdots$ & \\
	&Context 10:& \makecell[l]{Carson subtly defers her writing in just the right\\ writing for it to not be subject to an induction run\\ rampant style which grabs the readers interest without\\ biasing the whole article.}  \\
	\cmidrule{2-3}
	& Question: & Why did Rachel Carson write an obligation to endure?\\
	\cmidrule{2-3}
	& Answer: & \makecell[l]{Rachel Carson writes The Obligation to Endure\\ because believes that as man tries to eliminate\\ unwanted insects and weeds; however he is actually\\ causing more problems by polluting the environment.} \\
	\noalign{\hrule height 1pt}

\end{longtable}

\subsection{Comparison of Different Tasks}
To compare the contribution and limitations of the four MRC tasks, we evaluated five dimensions: construction, understanding, flexibility, evaluation and application. In each dimension, we compare tasks and score them according to relative ranking. As there are four tasks under consideration, score 4 means the corresponding task has the best performance in that dimension while score 1 means it performs the worst. Moreover, two tasks score similarly means that it is hard to judge which one is better in the dimension.
\begin{itemize}[leftmargin=*,labelsep=5.8mm]
	\item[-] Construction: This dimension measures whether it is easy to construct datasets for the task or not. The easier it is, the higher the score.
	\item[-] Understanding: This dimension evaluates how well the task can test the machine's ability to understand. If a task needs more understanding and reasoning, the score is higher.
	\item[-] Flexibility: The flexibility of the answer form can measure the quality of the tasks. When answers are more flexible, the flexibility score is higher.
	\item[-] Evaluation: Evaluation is a necessary part of MRC tasks. Whether a task can be easily evaluated also determines its quality. Tasks that are easy to evaluate get high scores in this dimension.
	\item[-] Application: A good task is supposed to be close to real-world application. Therefore, scores in this dimension are high, if a task can easily be applied to the real world.
\end{itemize}

\begin{figure}[!h]
	\centering
	\includegraphics[width=8.5cm]{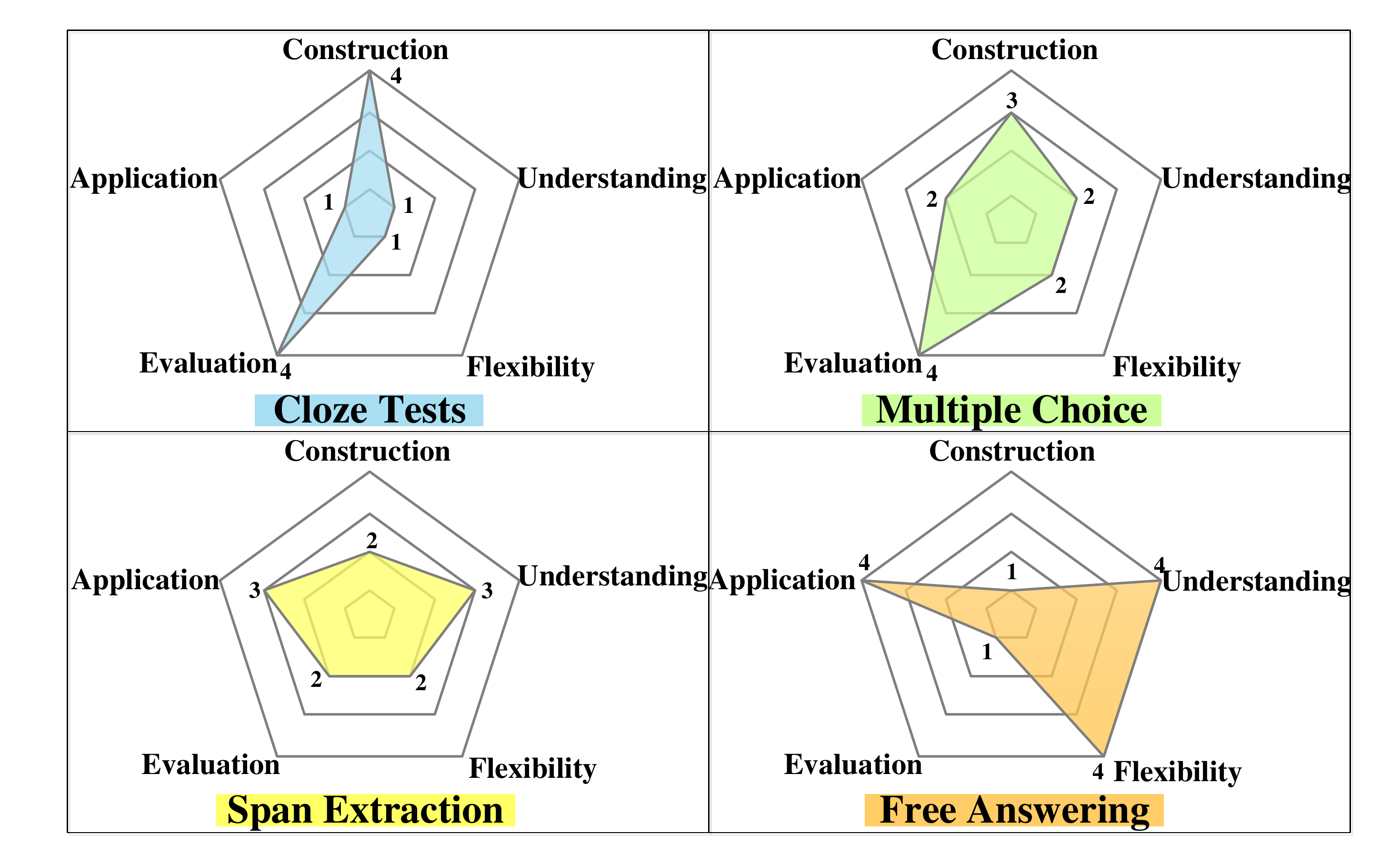}
	\caption{Comparison of different MRC tasks.}
	\label{compare}
\end{figure}

As presented in Figure \ref{compare}, scores of these five dimensions vary from different tasks. \mbox{More specifically}, it is easiest to construct datasets and evaluate cloze tests. However, as the answer form is restricted to a single word or name entity in the original context, cloze tests cannot test machine understanding well and do not conform with real-world applications. Multiple-choice tasks provide candidate answers for each question, so that even if answers are not limited in the original context, they can be easily evaluated. It is not very hard to build datasets for this task as multiple-choice tests in language exams can be easily used. However, candidate answers lead to a gap between synthetic datasets and realistic application. In contrast, span extraction tasks are a moderate choice, for which datasets can easily be constructed and evaluated. Moreover, they can test a machine's understanding of text, in a way. All these advantages contribute to quite a lot research focusing on these tasks. The disadvantage of span extraction is that answers are constrained to the subsequence of original context, which is still a little far from the real world. Free answering tasks show their superiority in the dimensions of understanding, flexibility, and application, which are the closest to practical application. However, every coin has two sides. Because of the flexibility of the answer form, it is somewhat hard to build datasets, and how to effectively evaluate performance on these tasks remains a challenge.

\section{Deep-Learning-Based Methods} \label{s3}

\par With the release of the CNN \& Daily Mail dataset \cite{hermann2015teaching} and the development of deep-learning techniques, neural MRC shows superiority over traditional rule-based and machine-learning-based MRC, and has gradually become the mainstream in the research community. In this section, we~introduce the general architecture of neural MRC systems, followed by an illustration of typical deep-learning methods applied to different modules.

\subsection{General Architecture}
\begin{figure}[h]
	\centering
	\includegraphics[width=8.5cm]{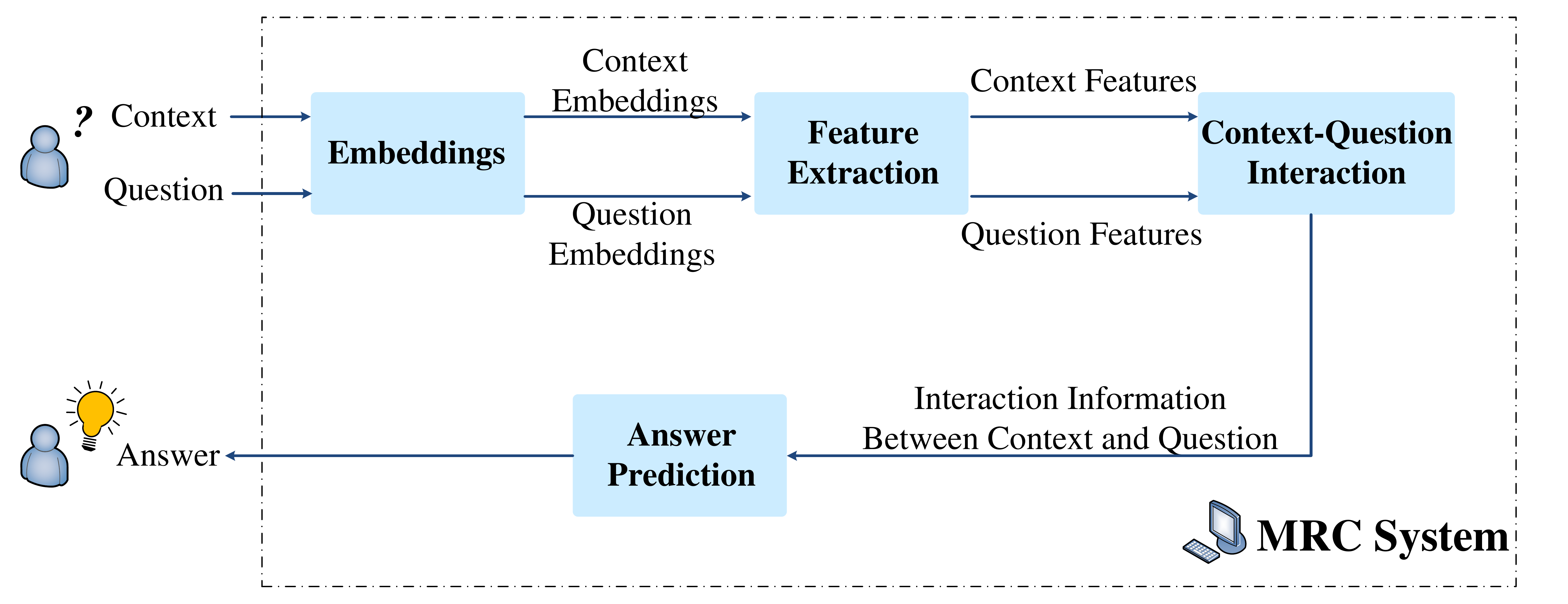}
	\caption{The general architecture of machine reading comprehension system.}
	\label{architecture}
\end{figure}

As presented in Figure \ref{architecture}, a typical neural machine reading comprehension system, which takes the context and question as inputs and the answer as output, contains four key modules: embeddings, feature extraction, context-question interaction and answer prediction. The function of each module can be interpreted as follows.

\begin{itemize}[leftmargin=*,labelsep=5.8mm]
	\item[-] Embeddings: As the machine is unable to understand natural language directly, the embedding module is indispensable to change input words into fixed-length vectors at the beginning of the MRC systems. Taking the context and question as inputs, this module outputs context and question embeddings by various approaches. Classical word representation methods such as one-hot and Word2Vec, sometimes combined with other linguistic features, i.e., part-of-speech, name entity, and question category, are usually used to represent semantic and syntactic information in the words. Moreover, contextualized word representations pre-trained by a large corpus also show promising performance in encoding contextual information.
	\item[-] Feature Extraction: After the embedding module, context and question embeddings are fed to the feature extraction module. To better understand the context and question, this module is aimed at extracting more contextual information. Some typical deep neural networks, such as recurrent neural networks (RNNs) and convolution neural networks (CNNs) are applied to further mine contextual features from context and question embeddings.
	\item[-] Context-Question Interaction: The correlation between the context and the question plays a significant role in predicting the answer. With such information, the machine can find out which parts in the context are more important to answering the question. To achieve that goal, the attention mechanism, unidirectional or bidirectional, is widely used in this module to emphasize parts of the context relevant to the query. To sufficiently extract their correlation, the interaction between the context and the question sometimes involves multiple hops, which simulates the rereading process of human comprehension.
	\item[-] Answer Prediction: Answer prediction module, the last component of an MRC system, outputs the final answer based on all the information accumulated from previous modules. As MRC tasks can be categorized according to answer forms, this module is highly related to different tasks. For cloze tests, the output of this module is a word or entity in the original context, while the multiple-choice task requires selecting the correct answer from candidate answers. In terms of span extraction, this module extracts a subsequence of the given context as the answer. Some generation techniques are used in this module for the free answering task,  as it has nearly no constraint on answer forms. 
\end{itemize}

\subsection{Typical Deep-Learning Methods}
\begin{figure}[h]
	\centering
	\includegraphics[width=10cm]{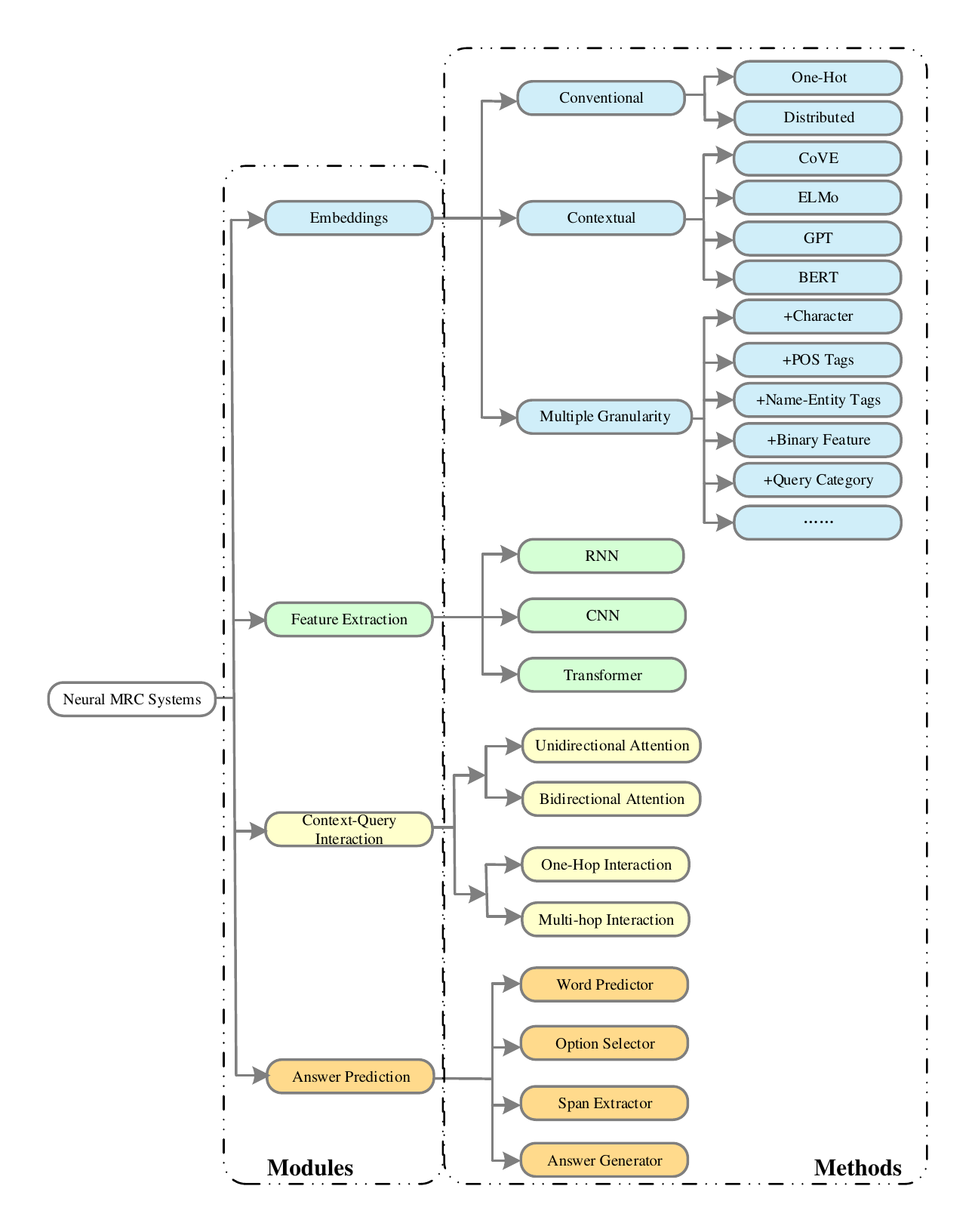}
	\caption{Typical techniques in neural MRC systems.}
	\label{method}
\end{figure}

Compared to traditional rule-based and machine-learning-based methods, deep-learning techniques show their superiority in extracting contextual information, which is very important to MRC tasks. In this section, we present various deep-learning approaches used in different modules of MRC systems in Figure \ref{method} and summarize those approaches applied in classic MRC models in Table \ref{classic}. Moreover, some tricks to improve the performance of MRC systems are introduced in the last part of this section.

\begin{table}[h]
	\centering
	\renewcommand{\arraystretch}{1.2}
	\caption{\upshape{Classic neural MRC models(C: Cloze Tests, M: Multiple Choice, S: Span Extraction, F:~Free~Answering).}}
	\label{classic}
	\scalebox{.78}[.78]{\begin{tabular}{ccccc}
			\toprule
			\multirow{2}{*}{\bfseries{Models}}& \multicolumn{4}{c}{\bfseries{Methods}}\\
			\cmidrule{2-5}
			& \bfseries{Embeddings} & \makecell[c]{\bfseries{Feature}\\ \bfseries{Extraction}} & \makecell[c]{\bfseries{Context-Question} \\ \bfseries{Interaction}} & \makecell[c]{\bfseries{Answer}\\  \bfseries{Prediction}}\\
			\midrule
			Attentive Reader (Hermann et al. 2015) \cite{hermann2015teaching}&Conventional &RNN &\makecell[c]{Unidirectional\\One-Hop}& \makecell[c]{Word\\ Predictor}\\
			\midrule
			Impatient Reader (Hermann et al. 2015) \cite{hermann2015teaching}&Conventional &RNN &\makecell[c]{Unidirectional\\Multi-hop} & \makecell[c]{Word\\ Predictor}\\
			\midrule
			\makecell[l]{End-to-End Memory Networks\\ (Sukhbaatar et al. 2015) \cite{sukhbaatar2015end}}&Conventional&/&Multi-Hop&\makecell[c]{Word\\ Predictor}\\
			\midrule
			Standford Reader (Chen et al. 2016) \cite{chen2016thorough} &Conventional &RNN &\makecell[c]{Unidirectional\\One-Hop} &\makecell[c]{Word\\ Predictor}\\
			\midrule
			AS Reader (Kadlec et al. 2016) \cite{kadlec2016text} &Conventional &RNN &\makecell[c]{Unidirectional\\One-Hop}&\makecell[c]{Word\\ Predictor}\\
			\midrule
			IA Reader (Sordoni et al. 2016) \cite{sordoni2016iterative} &Conventional &RNN &\makecell[c]{Unidirectional\\Multi-Hop}&\makecell[c]{Word\\ Predictor} \\
			\midrule
			Match-LSTM (Wang \& Jiang 2016) \cite{wang2016machine} &Conventional &RNN &\makecell[c]{Unidirectional\\Multi-Hop}&\makecell[c]{Span\\ Extractor} \\
			\midrule
			Bi-DAF (Seo et al. 2016) \cite{seo2016bidirectional} &\makecell{Multiple\\Granularity} &RNN &\makecell[c]{Bidirectional\\One-Hop}&\makecell[c]{Span\\ Extractor}\\
			\midrule
			DCN (Xiong et al. 2016) \cite{xiong2016dynamic} &Conventional &RNN &\makecell[c]{Bidirectional\\One-Hop}&\makecell[c]{Span\\ Extractor} \\
			\midrule
			AoA Reader (Cui et al. 2017) \cite{cui2017attention} &Conventional &RNN &\makecell[c]{Bidirectional\\One-hop}&\makecell[c]{Word\\ Predictor}\\
			\midrule
			GA Reader (Dhingta et al. 2017) \cite{dhingra2017gated}&Conventional &RNN &\makecell[c]{Unidirectional\\Multi-Hop}&\makecell[c]{Word\\ Predictor}\\
			\midrule
			MEMEN (Pan et al. 2017) \cite{pan2017memen}&\makecell[c]{Multiple\\Granularity}&RNN&Multi-Hop&\makecell[c]{Span\\ Extractor} \\
			\midrule
			R-NET (Wang et al. 2017) \cite{wang2017gated}&\makecell{Multiple\\Granularity} &RNN &\makecell[c]{Unidirectional\\Multi-Hop}&\makecell[c]{Span\\ Extractor} \\
			\midrule
			DCN+ (Xiong et al. 2017) \cite{xiong2017dcn+}&Conventional & RNN &\makecell[c]{Bidirectional\\Multi-Hop}&\makecell[c]{Span\\ Extractor} \\
			\midrule
			S-NET (Tan et al. 2017) \cite{tan2018s}&\makecell{Multiple\\Granularity}& RNN &\makecell[c]{Unidirectional\\Multi-Hop}&\makecell[c]{Answer\\Generator} \\
			\midrule
			CoVe + DCN (Mccann et al. 2017) \cite{mccann2017learned}& Contextual & RNN &\makecell[c]{Bidirectional\\One-Hop}&\makecell[c]{Span\\ Extractor}\\
			\midrule
			QANet (Yu et al. 2018) \cite{yu2018qanet}&\makecell{Multiple\\Granularity} &Transformer&\makecell[c]{Bidirectional\\One-Hop}&\makecell[c]{Span\\ Extractor}\\
			\midrule
			\makecell[l]{Reinforced Mnemonic Reader\\ (Hu et al. 2018) \cite{hu2018reinforced}}&\makecell{Multiple\\Granularity}&RNN &\makecell[c]{Bidirectional\\Multi-Hop}&\makecell[c]{Span\\ Extractor}\\
			\midrule
			CSA (Chen et al. 2018) \cite{chen2018convolutional}&\makecell{Multiple\\Granularity}&RNN\&CNN&\makecell[c]{Bidirectional\\One-Hop}&\makecell[c]{Option\\ Selector}\\
			\midrule
			CNN Model (Chaturvedi et al. 2018) \cite{chaturvedi2018cnn}&Conventional&CNN&\makecell[c]{Unidirectional\\One-Hop}&\makecell[c]{Option\\ Selector}\\
			\midrule
			\makecell[l]{Hierarchical Attention Flow\\ (Zhu et al. 2018) \cite{zhu2018hierarchical}}&Conventional&RNN&\makecell[c]{Bidirectional\\One-Hop}&\makecell[c]{Option\\ Selector}\\
			\midrule
			\makecell[l]{ELMo + improved Bi-DAF\\ (Peters et al. 2018) \cite{peters2018deep}}& Contextual & RNN &\makecell[c]{Bidirectional\\One-Hop}&\makecell[c]{ Span\\ Extractor}\\
			\midrule
			GPT (Radford et al. 2018)\cite{radford2018improving}& Contextual&/ &/ &\makecell[c]{Option\\ Selector}\\
			\midrule
			BERT (Devlin et al. 2018) \cite{devlin2019bert}& Contextual&/ &/ &\makecell[c]{Span\\ Extractor}\\
			\bottomrule
	\end{tabular}}
\end{table}

\subsubsection{Embeddings}

The embedding module is an essential part of an MRC systems and is usually placed at the beginning to encode input natural language words into fixed-length vectors, which the machine can understand and deal with. As Dhingra et al. \cite{dhingra2017comparative} point out, minor choices made in word representation can lead to substantial differences in the final performance of the reader. How to sufficiently encode the context and question is the pivotal task of this module. In existing MRC models, word representation methods can be sorted into conventional word representation and pre-trained contextualized representation. To encode more abundant semantic and linguistic information, multiple granularity, which fuses word-level embeddings with character-level embeddings, part-of-speech, name entity, word frequency, question category, and so on, is also applied to some MRC systems. \mbox{In the following} parts, we will give a detailed illustration.

\noindent(1) Conventional Word Representation	

\begin{itemize}[itemindent=2em,listparindent=2em,leftmargin=0pt,itemsep=10pt,parsep=10pt]
	\item[-] One-Hot
	\par This method \cite{baeza2011modern} represents words with binary vectors, and its size is same as the number of words in the dictionary. In such vectors, one position is 1, corresponding to the word, while all others are 0. As a word representation approach at the early stage, it can encode words when the size of the vocabulary is not very large. However, this representation is sparse and may suffer from the curse of dimensionality with increased vocabulary. In addition, one-hot encoding cannot represent relationships among words. For instance, ``apple'' and ``pear'' belong to the category ``fruit'', but their word representations embedded by one-hot cannot show such a relationship.  
	\newpage
	\item[-] Distributed Word Representation
	\par To address the shortcomings of representations such as one-hot, Rumelhart et al. \cite{rumelhart1986learning} propose distributed word representation, which encodes words into continuous low-dimensional vectors. Closely related words encoded by these methods are not far from each other in vector space, which reveals the correlation of words. Various techniques to generate distributed word representations have been introduced, among which the most popular are Word2Vec \cite{mikolov2013distributed} and GloVe \cite{pennington2014glove}. Besides successful applications in a variety of Natural Language Processing (NLP) tasks, such as machine \mbox{translation \cite{cho2014learning}} and sentiment analysis \cite{nakov2016semeval}, vectors produced by these methods are also applied to many MRC systems.
\end{itemize}

\noindent(2) Pre-trained Contextualized Word Representation
\par Although distributed word representation can encode words in low-dimensional space and reflect correlations between words, it cannot efficiently mine contextual information. To be specific, a vector produced by the distributed word representation of one word is constant regardless of different contexts. To address this problem, researchers have introduced contextualized word representations, which are pre-trained with large corpora in advance and then directly used as conventional word representations or fine-tuned according to the specific tasks. This is a kind of transfer learning and has shown promising performance in a wide range of NLP tasks including machine reading comprehension. Even a simple neural network model can perform well in answer prediction with such a pre-trained word representation approach.

\begin{itemize}[itemindent=2em,listparindent=2em,leftmargin=0pt,itemsep=10pt,parsep=10pt]
	\item[-] CoVE
	\par Inspired by successful cases in computer vision, which transfer CNNs pre-trained on a large supervised training corpus such as ImageNet to other tasks, McCann et al. \cite{mccann2017learned} try to bring the benefit of transfer learning to NLP tasks. They first train long short-term memory (LSTM) encoders of sequence-to-sequence models on a large-scale English-to-German translation dataset and then transfer the outputs of the encoder to other NLP tasks. As machine translation (MT) requires the model to encode words in context, the output of the encoder can be regarded as context vectors (CoVE). To deal with MRC problems, McCann et al. concatenate the outputs of an MT encoder with word embeddings pre-trained by GloVe to represent the context and question and feed them through the coattention and dynamic decoder implemented in a dynamic coattention network (DCN) \cite{xiong2016dynamic}. DCN with CoVe outperforms the original DCN on the SQuAD dataset, which illustrates the contribution of contextualized word representations to downstream tasks. However, pre-training CoVE requires a large parallel corpus. Its performance will degrade if the training corpus is not adequate.
	\item[-] ELMo
	\par \par Embeddings from language models (ELMo), proposed by Peters et al. \cite{peters2018deep}, is another type of contextualized word representation. To get ELMo embeddings, they first pre-train a bidirectional language model (biLM) with a large text corpus. Compared to CoVe, ELMo breaks the constraint of limited  parallel corpus and obtains richer word representations by collapsing outputs of all biLM layers into a single vector with a task-specific weighting rather than using outputs of the top layer. Model evaluations illustrate that different levels of LSTM states can capture diverse syntactic and linguistic information. When applying ELMo embeddings to MRC models, Peters et al. choose an improved version of bidirectional attention flow (Bi-DAF), introduced by Clark and Gardner \cite{clark2018simple}, as the baseline and improved the state-of-the-art single model by 1.4\% on the SQuAD dataset. ELMo, which can be easily integrated with existing models, shows promising performance on various NLP tasks, but it is limited by the insufficient feature extraction capability of LSTM.
	\item[-] GPT
	\par Generative pre-training (GPT) \cite{radford2018improving}, is a semi-supervised approach combining unsupervised pre-training and supervised fine-tuning. Representations pre-trained by this method can transfer to various NLP tasks with little adaptation. The basic component of GPT is a multi-layer transformer~\cite{vaswani2017attention} decoder that mainly uses multi-head self-attention to train the language model and allow longer semantic structure to be captured compared to RNN-based models. After training, the pre-trained parameters are fine-tuned for specific downstream tasks. In terms of MRC problems such as multiple choice, Radford et al. concatenate the context and the question with each possible answer and process such sequences with transformer networks. Finally, they produce an output distribution over possible answers to predict correct answers. GPT achieves an improvement of 5.7\% on the RACE~\cite{lai2017race} dataset compared with the state-of-the-art. Seeing the benefit brought by contextualized word representations pre-trained on large-scale datasets, Radford et al. \cite{radford2019language} later propose GPT-2, which is pre-trained on a larger corpus, WebText, with more than 1.5 billion parameters. Compared to the previous model, the layers of transformer architecture are increased from 12 to 48. Moreover, single task training is substituted with a multi-task learning framework, which makes GPT-2 more generative. \mbox{This improved} version shows competitive performance even in zero-shot setting. However, the transformer architecture used in GPT and GPT-2 is unidirectional (left-to-right), which cannot incorporate  context from both directions. This may be the major shortcoming and limits its performance on downstream tasks.
	\item[-] BERT
	\par Considering the limitations of unidirectional architecture applied in previous pre-training models such as GPT, Devlin et al. \cite{devlin2019bert} propose a new model, bidirectional encoder representation from transformers (BERT). With the masked language model (MLM) and next-sentence prediction task, BERT can pre-train deep contextualized representations with a bidirectional transformer, encoding both left and right context for word representations. As transformer architecture cannot extract sequential information, Devlin et al. add positional embeddings to encode position. Owing to the bidirectional language model and transformer architecture, BERT outperforms state-of-the-art models on 11 NLP tasks. In particular, for MRC tasks, BERT is so competitive that using it with a simple answer prediction approach shows promise. Despite its outstanding performance, BERT's pre-training process is time and resource-consuming which makes it nearly impossible to pre-train without abundant computational resources.
\end{itemize}

\noindent(3) Multiple Granularity

\par Word-level embeddings pre-trained by Word2Vec or GloVe cannot encode rich syntactic and linguistic information, such as part-of-speech, affixes, and grammar, which may not be sufficient for deep machine understanding. To incorporate fine-grained semantic information into word representations, some researchers have introduced approaches to encode the context and the question at different levels of granularity.
\begin{itemize}[itemindent=2em,listparindent=2em,leftmargin=0pt,itemsep=10pt,parsep=10pt]
	\item[-] Character Embeddings
	\par Character embeddings represent words at the character level. Compared to word-level representations, they are not only more suitable for modeling sub-word morphologies but also can alleviate the out-of-vocabulary (OOV) problem. Seo et al. \cite{seo2016bidirectional} add character-level embeddings to their Bi-DAF model for the MRC task. They use CNNs to obtain character-level embeddings. \mbox{Each character} in the word is embedded into a fixed-dimension vector, which is fed to CNNs as 1D inputs. \mbox{After max-pooling} the entire width, the outputs of CNNs are embeddings at the character level. \mbox{The concatenation} of word-level and character-level embeddings are then fed to the next module as input. In addition, character embeddings can be encoded with bidirectional LSTMs \cite{hu2018reinforced,wang2016multi}. \mbox{For each} word, the outputs of the last hidden state are considered to be its character-level representation. Moreover, word-level and character-level embeddings can be combined dynamically with a fine-grained gating mechanism rather than simple concatenation to mitigate the imbalance between frequent and infrequent words \cite{yang2016words}.
	\item[-] Part-of-Speech Tags
	\par A part-of-speech (POS) is a particular grammatical class of words, such as nouns, adjectives, or verb. Labeling POS tags in NLP tasks can illustrate complex characteristics of word use and in turn contribute to disambiguation. To translate POS tags into fixed-length vectors, they are regarded as variables, randomly initialized in the beginning and updated while training. 
	\item[-] Name-Entity Tags
	\par Name entity, a concept in information retrieval, refers to a real-world object, such as a person, location, or organizations, with a proper name. When asking about such objects, name entities are probable answer candidates. Thus, embedding name-entity tags of context words can improve the accuracy of answer prediction. The method of encoding name-entity tags is similar to that of POS tags mentioned above.
	\item[-] Binary Feature of Exact Match (EM)
	\par This feature, which measures whether a context word is in the question, was first used in the conventional entity-centric model proposed by Chen et al. \cite{chen2017reading}. Later, some researchers used it in the Embedding module to enrich word representations. The value of this binary is 1 if a context word can be exactly matched to one word in the query, otherwise its value is 0. More loosely, Chen~et~al.~\cite{chen2018convolutional} use partial matching to measure the correlation between context words and question words. \mbox{For instance,} ``teacher'' can be partially matched with ``teach''. 
	\item[-] Query-Category
	\par The types of questions (what, where, who, when, why, how) can usually provide clues to search for the answer. For instance, a ``where'' question pays more attention to spatial information. \mbox{Zhang et al. \cite{zhang2017exploring}} introduce a method to model different question categories in end-to-end training. They first obtain query types by counting the key word frequency. Then the question type information is encoded to one-hot vectors and stored in a table. For each query, they look up the table and use a feed-forward neural network for projection. The query-category embeddings are often added to the query word embeddings.
\end{itemize}
\vspace{-8pt}
\par The embeddings introduced above can be combined freely in the embedding module. \mbox{Hu et al.} \cite{hu2018reinforced} use word-level, character-level, POS tags, name-entity tags, binary feature of EM, and query-category embeddings in their Reinforced Mnemonic Reader to incorporate syntactic and linguistic information in word representations. Experimental results show that rich word representations contribute to deep understanding and improve answer prediction accuracy. 

\par To sum up, word embeddings encoded by distributed word representations are the basis of this module. As more abundant representations with syntactic and linguistic information contribute to better performance, multiple granularity representations have gradually become prevalent. In terms of contextual word representation, they can improve performance dramatically, and can be used by themselves or combined with other representations.

\subsubsection{Feature Extraction}\label{my_section}

The feature extraction module is often placed after the embedding layer to extract features of the context and question separately. It further pays attention to mining contextual information at the sentence level based on various types of syntactic and linguistic information encoded by the embedding module. Recurrent neural networks (RNNs), convolution neural networks (CNNs) and Transformer architecture are applied in this module, and we will give an illustration in detail \mbox{in this part.}

\noindent(1) Recurrent Neural Networks
\par RNNs are popular models that have shown great promise for dealing with sequential information. RNNs are called \textit{recurrent} as outputs in each time step depending on the previous computations. RNN-based models have been widely used in various NLP tasks, such as machine translation, sequence tagging and question answering. In particular, long short-term memory (LSTM) \cite{hochreiter1997long} networks and gated recurrent units (GRUs) \cite{cho2014learning}, variants of RNNs, are much better at capturing long-term dependencies than plain ones are and can alleviate gradient explosion and vanishing problems. Since the preceding and following words have the same importance in understanding the given word, many researchers use bidirectional RNNs to encode the context and question embeddings in MRC systems. The context and question embeddings are denoted as $x_{p}$ and $x_{q}$, respectively, and we will illustrate how the feature extraction module with bidirectional RNNs handles those embeddings and extracts sequential information.
\par In terms of questions, the feature extraction process with bidirectional RNNs can be sorted into two types: word-level and sentence-level. 
\par In word-level encoding, feature extraction outputs for each question embedding $x_{qj}$ at time step $j$ can be denoted as follows:

\begin{figure}[h]
	\centering
	\includegraphics[width=11cm]{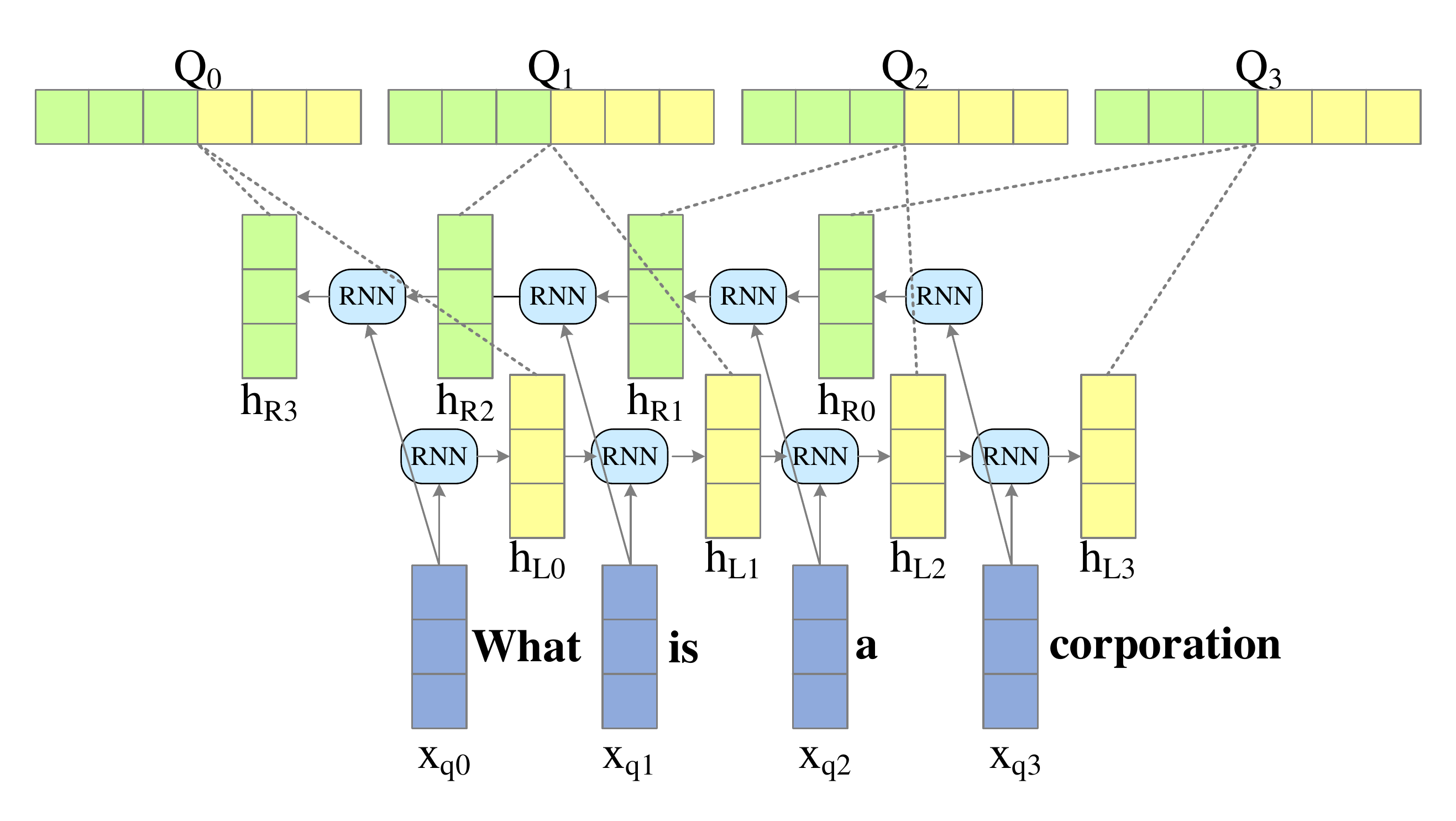}
	\caption{Word-level encoding for questions.}
	\label{question1}
\end{figure}

\begin{equation}
Q_{j} = [\overrightarrow{RNN}(x_{qj});\overleftarrow{RNN}(x_{qj})],
\end{equation}
where $\overrightarrow{RNN}(q_{xj})$ and $\overleftarrow{RNN}(q_{xj})$ denotes forward and backward hidden states of bidirectional RNNs, respectively. This process is shown in Figure \ref{question1} in detail.

By contrast, sentence-level encoding regards the question sentence as a whole. The feature extraction process can be denoted as:

\begin{figure}[h]
	\centering
	\includegraphics[width=9.5cm]{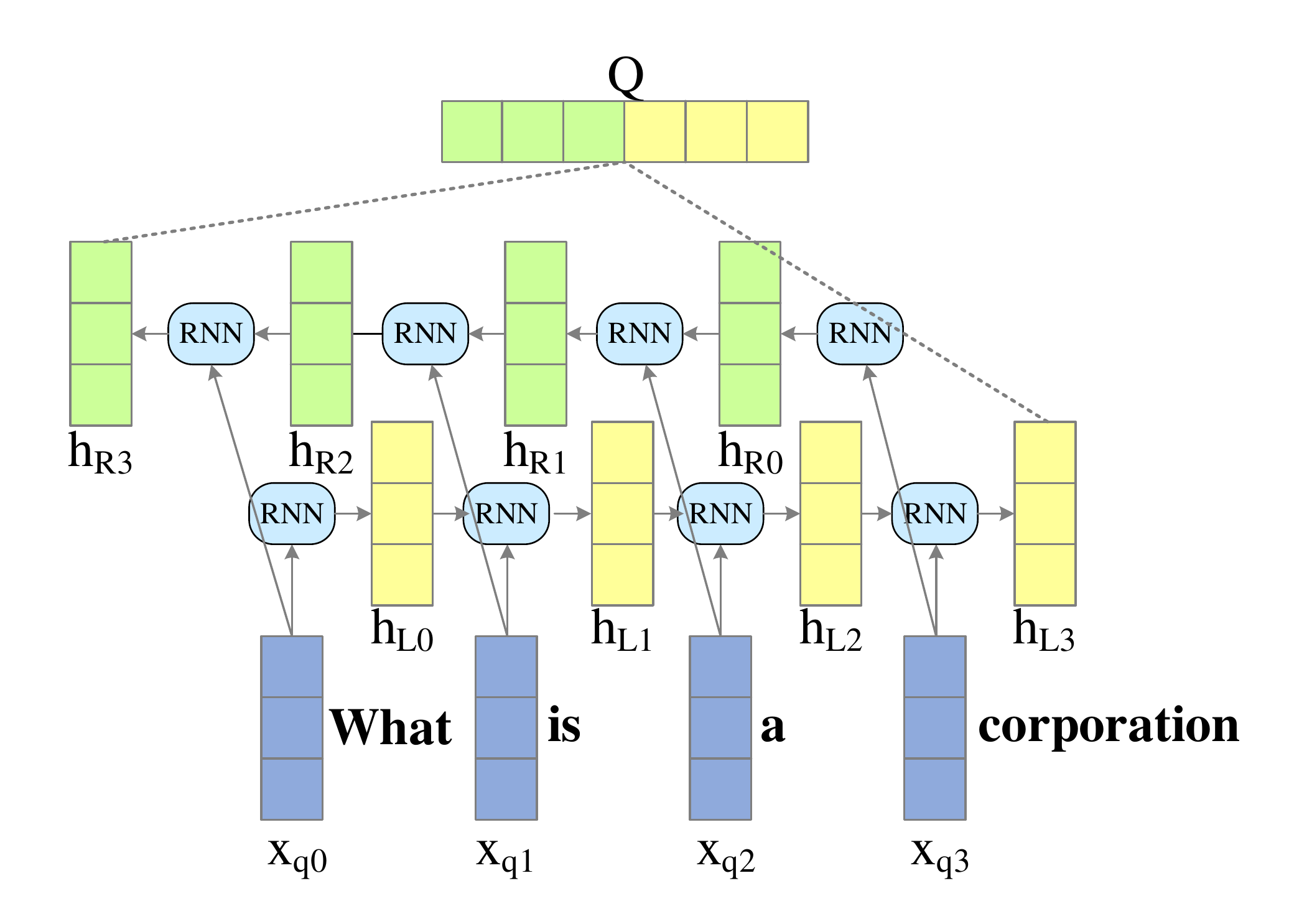}
	\caption{Sentence-level encoding for questions.}
	\label{question2}
\end{figure}

\begin{equation}
Q = [\overrightarrow{RNN}(x_{q|l|});\overleftarrow{RNN}(x_{q0})],
\end{equation}
where $|l|$ is the length of the question, $\overrightarrow{RNN}(x_{q|l|})$ and $\overleftarrow{RNN}(x_{q0})$ represent final forward and backward outputs of RNNs, respectively. To be more concrete, we demonstrate this sentence-level encoding process in Figure \ref{question2}.

\par As the context in an MRC task is usually a long sequence, researchers use word-level feature extraction to encode sequential information of context. Similar to question encoding, the feature extraction process with bidirectional RNNs for context embedding $x_{pi}$ at time step $i$ can be denoted as:

\begin{equation}
P_{i} = [\overrightarrow{RNN}(x_{pi});\overleftarrow{RNN}(x_{pi})].
\end{equation}

Although RNNs are capable of modeling sequential information, their training process is time-consuming as they cannot be processed in parallel.

\noindent(2) Convolution Neural Networks
\par
\begin{figure}[h]
	\centering
	\includegraphics[width=0.8\textwidth]{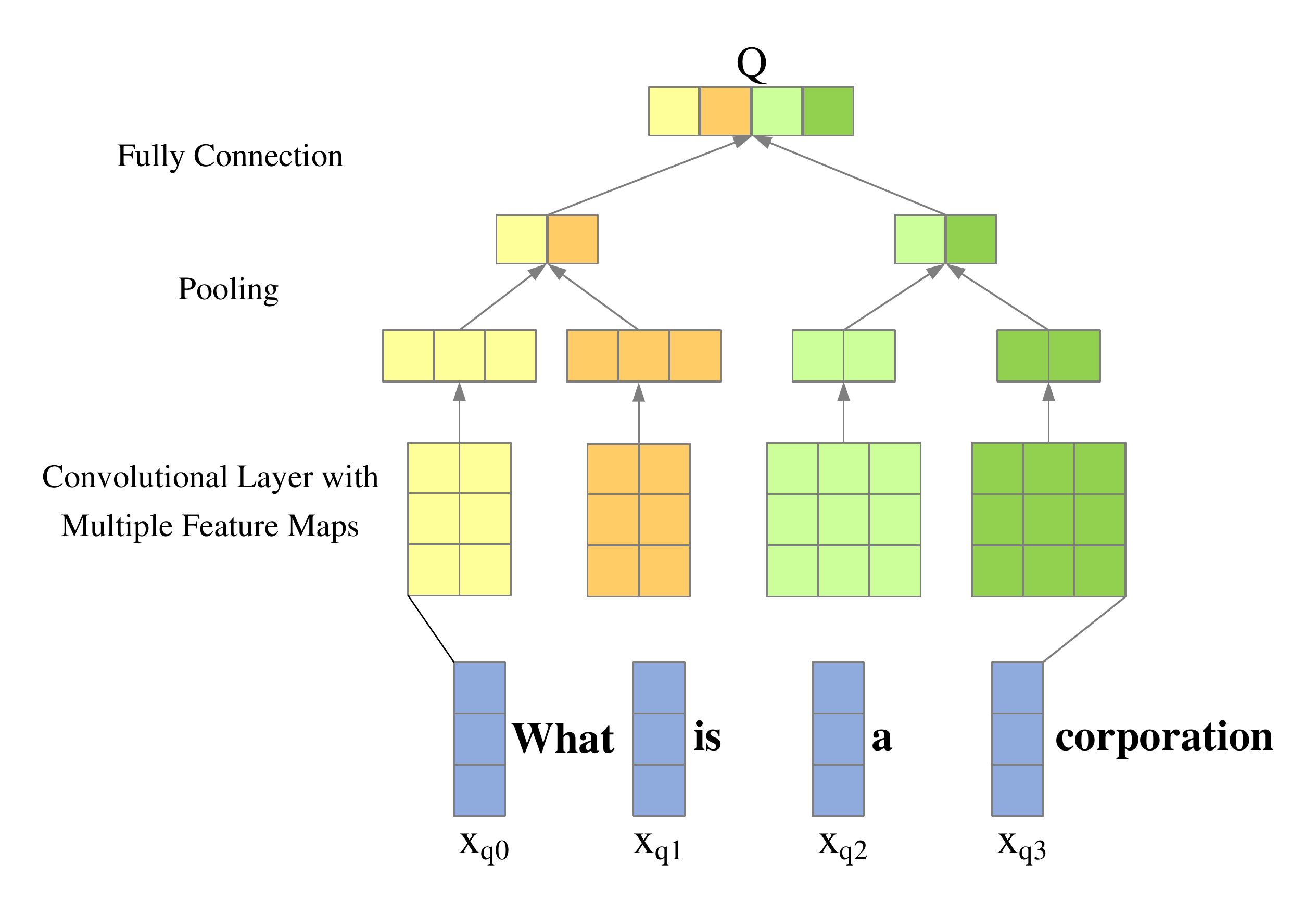}
	\caption{Using CNNs to extract features of question.}
	\label{cnn}
\end{figure}

 CNNs are widely used in computer vision. When applied to NLP tasks, one-dimensional CNNs show their superiority in mining local contextual information with sliding windows. In CNNs, each convolutional layer applies a different scale of feature maps to extract local features in diverse window sizes. The outputs are then fed to pooling layers to reduce dimensionality but keep the most significant information to the greatest extent. Maximum and average operations on the results of each filter are common ways to do pooling. Figure \ref{cnn} shows how the feature extraction module uses CNNs to mine the local contextual information of a question.

\par As shown in Figure \ref{cnn}, given word embeddings of a question $x_{q} \in \mathbb{R} ^{|l|\times d}$, where $|l|$ represents the length of the  question and $d$ denotes the dimension of word embeddings, the convolution layer has two filter sizes $f_{t} \times d (\forall t=2,3)$ with $k$ output channels ($k=2$ in the example in Figure \ref{cnn}). Each filter produces a feature map of shape $(|l|-t+1) \times k$ (padding is 0 and stride is 1), which are pooled to generate a k-dimensional vector. The two k-dimensional vectors are concatenated to form \mbox{a 2k-dimensional} vector, represented by $Q$.

\par Although both n-gram models and CNNs can focus on local features of the sentence, training parameters in n-gram models increase exponentially with larger vocabularies. By contrast, CNNs can extract local information in a more compact and efficient way regardless of the vocabulary size, as there is no need for them to represent every n-gram in the vocabulary. In addition, CNNs can be trained in parallel, which is faster than RNNs. One major shortcoming of CNNs is that they can extract only local information but are not capable of dealing with long sequence.

\noindent(3) Transformer 

\par
\begin{figure}[h]
	\centering
	\includegraphics[height=0.6\textwidth]{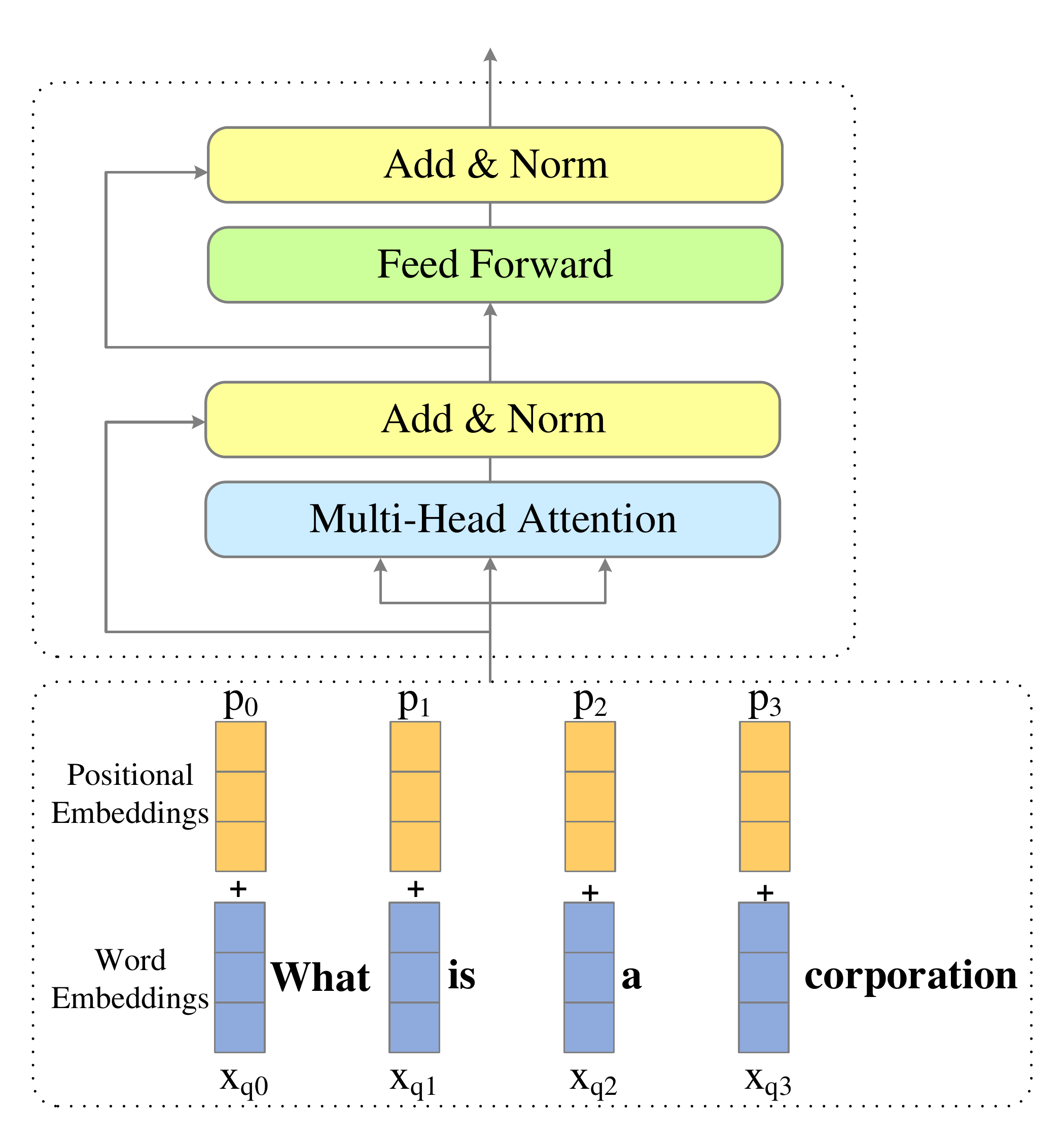}
	\caption{Using the Transformer to extract features of question.}
	\label{transformer}
\end{figure}

 The transformer, proposed by Vaswani et al. \cite{vaswani2017attention} in 2017, is a powerful neural network model that has shown promising performance on various NLP tasks \cite{radford2018improving,devlin2019bert}. In contrast to RNN-based or CNN-based models, the transformer is mainly based on the attention mechanism with neither recurrence nor convolution. Owing to multi-head self-attention, this simple architecture not only excels in alignment but also can be run in parallel. Compared to RNNs, the transformer requires less time to train, while it pays more attention to global dependencies, in contrast to CNNs. However, without recurrence and convolution, the model cannot make use of the order of the sequence. \mbox{To incorporate} positional information, Vaswani et al. add position encoding computed by sine and cosine functions. The sum of positional and word embeddings is fed to the transformer as inputs. Figure \ref{transformer} presents the simple architecture of the transformer. In practice, models usually stack several blocks with multi-head self-attention and feed-forward network. 

\par QANet, introduced by Yu et al. \cite{yu2018qanet}, is a representative MRC model that uses the transformer. \mbox{The basic} encoder block of QANet is a novel architecture, which combines the multi-head self-attention defined in the transformer with convolutions. The experimental results show that QANet achieves the same accuracy on SQuAD as the prevalent recurrent models with much faster training and inference~speed.

\par In general, most MRC systems use RNNs in the feature extraction module because of their superiority in handling sequential information. To accelerate the training process, some researchers substitute RNNs with CNNs or the transformer. CNNs are highly parallelized and can obtain rich local information with feature maps in different sizes. The transformer can mitigate the side-effect of long dependency and improve computational efficiency.

\subsubsection{Context-Question Interaction}

By extracting the correlation between the context and the question, models can find evidence for answer prediction. Inspired by Hu et al. \cite{hu2018reinforced}, we can divide existing works into two kinds according to how models extract correlations: one-hop and multi-hop interaction. No matter what kind of interaction the MRC model uses, the attention mechanism plays a critical role in emphasizing which parts of the context are more important to answer the questions.
\par Derived from human intuition, the attention mechanism was adapted for machine translation and shows promising performance on automatic token alignment \cite{bahdanau2014neural,luong2015effective}. As a simple and effective method that can be used to encode sequence data with its importance, it has shown significant improvement in various tasks in natural language processing, including text summarization \cite{rush2015neural}, sentiment classification \cite{wang2016attention}, semantic parsing \cite{cheng2016bi}, etc. In terms of machine reading comprehension, the attention mechanism can be categorized into unidirectional and bidirectional attention according to whether it is used unidirectionally or bidirectionally. In the following part, we introduce methods categorized using the attention mechanism, followed by illustrations of one-hop and \mbox{multi-hop interactions.}

\noindent(1) Unidirectional Attention

\par
\begin{figure}[h]
	\centering
	\includegraphics[width=0.9\textwidth]{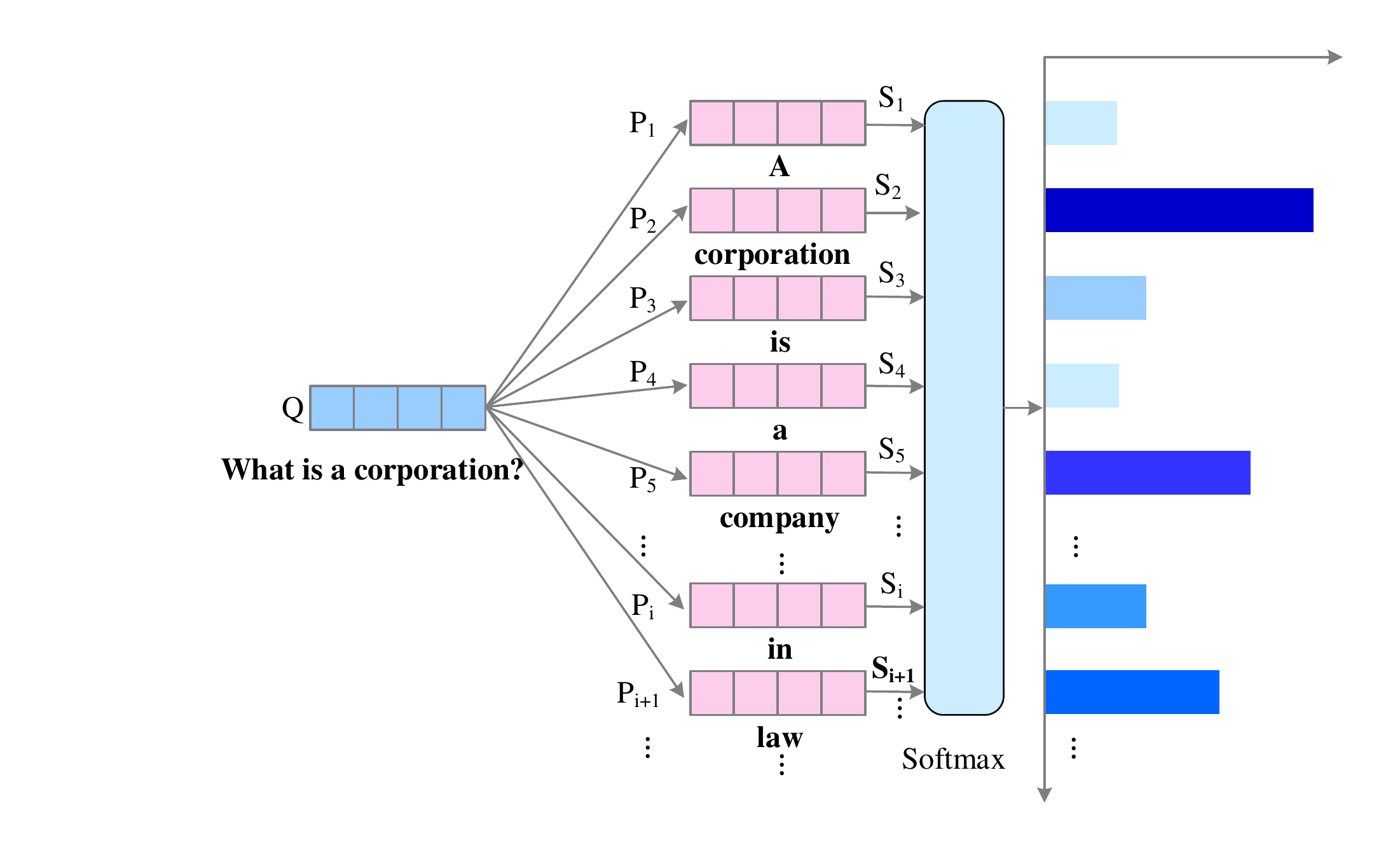}
	\caption{Using unidirectional attention to mine correlation between the context and question.}
	\label{unidirectional}
\end{figure}

 Unidirectional attention flow is usually from query to context, highlighting the most relevant parts of the context according to the question. It is believed that if the context word  is more similar to the question, it is more likely to be the answer word. As shown in Figure \ref{unidirectional}, the similarity of each context semantic embedding $P_{i}$ and the whole question sentence representations $Q$ (by sentence-level encoding introduced in Section \ref{my_section}) is calculated by $S_{i}=f(P_{i},Q)$, where $f(\cdot)$ represents the function that can measure the similarity. After being normalized by the SoftMax function in Equation (\ref{softmax}), \mbox{the attention} weight $\alpha_{i}$ for each context word is obtained, with which the MRC systems can finally predict the~answer.

\begin{equation}
\alpha_{i} = \frac{\text{exp}S_{i}}{\sum_{j}^{}\text{exp}S_{j}} \label{softmax}.
\end{equation}

\par The choice of function $f(\cdot)$ differs in different models. 
\par In the Attentive Reader proposed by Hermann et al. \cite{hermann2015teaching},   a tanh layer is used to compute \mbox{the relevance} between the context and the question as follows:

\begin{equation}
S_{i} = \text{tanh}(W_{P}P_{i}+W_{Q}Q),
\end{equation}
where $W_{P}$ and $W_{Q}$ are trainable parameters.
\par Following the work of Hermann et al., Chen et al. \cite{chen2016thorough} substitute a bilinear term for the tanh function as Equation (\ref{chen}): 

\begin{equation}
S_{i} = Q^{T}W_{s}P_{i} \label{chen},
\end{equation}
which makes the model simpler and more effective than the Attentive Reader.

\par The unidirectional attention mechanism can highlight the most important context words to answer the question. However, this method fails to pay attention to question words that are also pivotal for answer prediction. Hence, unidirectional attention flow is insufficient for extracting mutual information between the context and the query.

\noindent (2) Bidirectional Attention
\par Seeing the limitations of the unidirectional attention mechanism, some researchers introduced bidirectional attention flows, which not only computes query-to-context attention but also the reverse, context-to-query attention. This method, which takes a mutual look from both directions, can benefit from the interaction between context and query and provide complementary information.
\par
\begin{figure}[h]
	\centering
	\includegraphics[height=0.7\textwidth]{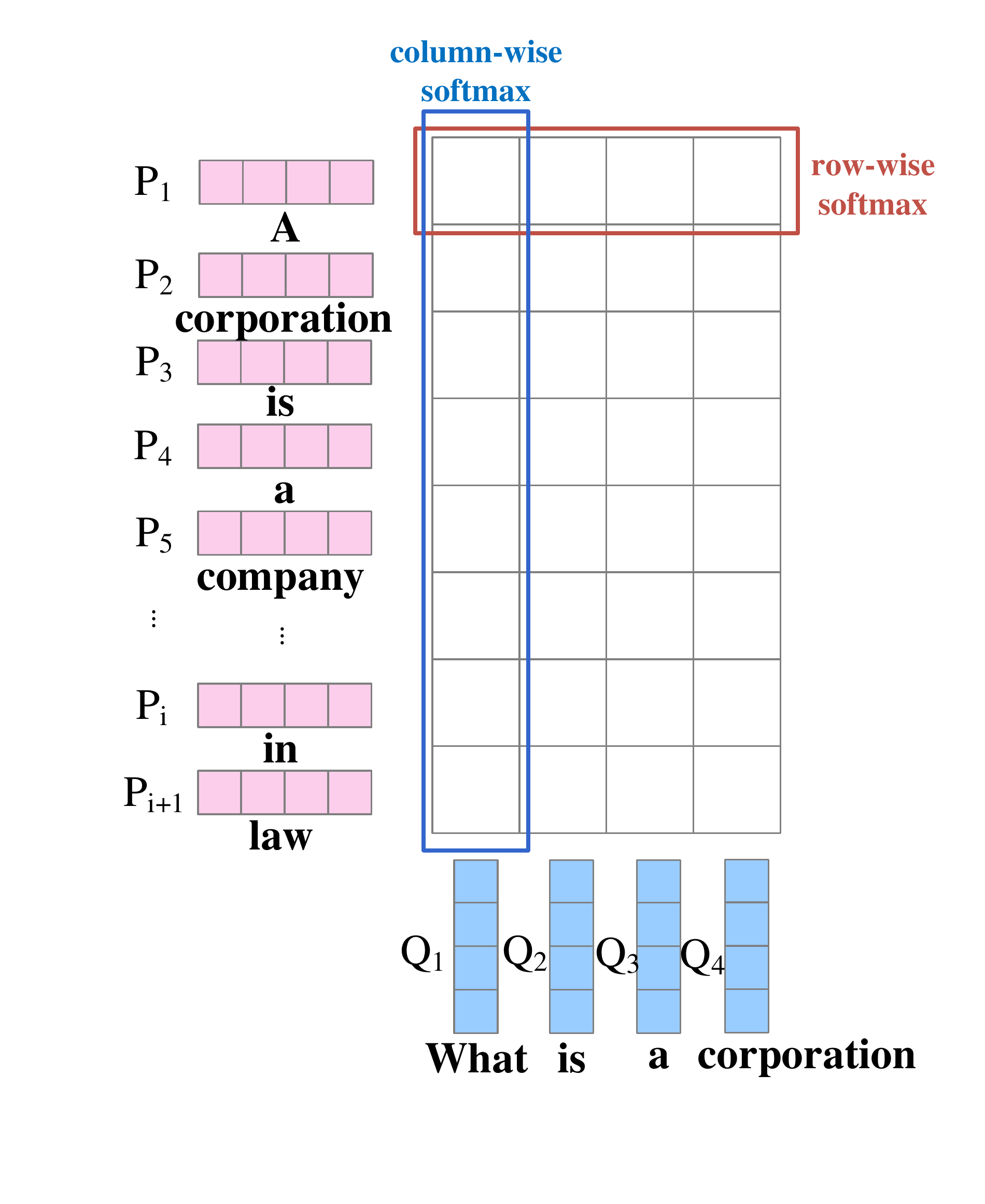}
	\caption{Using bidirectional attention to mine correlation between the context and question.}
	\label{bidirectional}
\end{figure}

 Figure \ref{bidirectional} presents the process of computing bidirectional attention. First, the pair-wise matching matrix $M(i,j)$ is obtained by computing the matching scores between each context semantic embedding $P_{i}$ and question semantic embedding $Q_{j}$ (by word-level encoding introduced in Section \ref{my_section}). \mbox{Then the outputs} of the column-wise SoftMax function can be regarded as query-to-context attention weight $\alpha$ while context-to-query attention $\beta$ is calculated by the row-wise SoftMax function.

\par The attention-over-attention reader (AoA Reader) model, the dynamic coattention network (DCN) and the bidirectional attention flow (Bi-DAF) network are representative MRC models with bidirectional attention.
\par In the AoA Reader, Cui et al. \cite{cui2017attention} compute the dot product between each context embedding and query embedding to obtain a similarity matching matrix $M(i,j)$. The query-to-context and context-to-query attention are calculated as presented in Figure \ref{bidirectional}. To combine these two attentions, different from previous work in CAS Reader \cite{cui2016consensus} using naive heuristics such as sum and average over the query-to-context attention, Cui et al. introduce attended attention, computed by the dot product of $ \alpha $ and the average result of $ \beta $, which is later used to predict the answer. 
\par To attend to the question and the document simultaneously, Xiong et al. \cite{xiong2016dynamic} fuse the two directional attentions as follows:

\begin{equation}
C = \alpha[Q;\beta P],
\end{equation}
where $C$ can be regarded as the coattention representations that contains attention information of both context and question. Based on DCN, Xiong et al. \cite{xiong2017dcn+} introduce an extension, DCN+, using residual connections to merge coattention outputs to encode richer information for the input sequences. Compared to the work on AoA Reader, Xiong et al. further calculate context representations with two types of directional attentive information, rather than directly using attention weights for answer prediction, which can better extract the correlation between context and question.

\par In contrast to the AoA Reader and DCN, which directly summarize the output of two directional attention flows, Seo et al. \cite{seo2016bidirectional} let attentive vectors flow into another RNN layer to encode query-aware context representations, which can reduce information loss caused by early summarization. \mbox{More specifically,} after obtaining query-to-context attention weight $\alpha$ and context-to-query attention weight $\beta$, Seo et al. compute attended context vector $\widetilde{P}$ and attended query vector $\widetilde{Q}$ as follows:

\begin{equation}
\begin{split}
&\widetilde{P} = \sum_{i}^{} \alpha P_{i},\\
&\widetilde{Q} = \sum_{j}^{} \beta Q_{j}.
\end{split}
\end{equation}

Then the context embeddings and attention vectors are combined by a simple concatenation:

\begin{equation}
G = [P;\widetilde{Q};P \circ \widetilde{Q};P \circ \widetilde{P}],
\end{equation}
where $\circ$ is element-wise multiplication and $G$ can be regarded as query-aware context representations, which are later fed to bidirectional LSTM to be further encoded.

To sum up, MRC systems at the early stage usually used unidirectional attention mechanism, especially query-to-context attention, to highlight the part of the context that was more important to answer the question. However, query-to-context attention is not sufficient to extract the mutual information between the context and the query. Later, bidirectional attention was widely applied to overcome the shortcomings of unidirectional attention, which benefited from context-query correlation and output attentive representations with a fusion of context and question information.

\noindent (3) One-Hop Interaction

\par One-hop interaction is a shallow architecture, where the interaction between the context and the question is computed only once. Earlier context-query interaction was such a one-hop architecture in many MRC systems; for example, the Attentive Reader \cite{hermann2015teaching}, the attention sum (AS) Reader \cite{kadlec2016text}, \mbox{the AoA} Reader \cite{cui2017attention} and so on. Although this method can do well in tackling simple cloze tests, when the question requires reasoning over multiple sentences in the context, it is hard for this one-hop interaction approach to predict the correct answer.

\noindent (4) Multi-Hop Interaction

\par In contrast to one-hop interaction, multi-hop interaction is much more complex; it tries to mimic the rereading phenomenon of humans by computing the interaction between the context and the question more than once. In the process of the interaction, whether information of the previous state, i.e., the memory of the already read context and question, can be efficiently stored or not directly affects the performance of the next interaction. 
\par There are mainly three methods to perform multi-hop interaction:
\par The first method calculates the similarity between the context and the question based on previous attentive representations of context. In the Impatient Reader model proposed by Hermann et al. \cite{hermann2015teaching}, query-aware context representations are dynamically updated by this method as each query token is read. This simulates the process of humans rereading a given context with the question information.

\par The second method introduces external memory slots to store previous memories. \mbox{The representative} model using this method is memory networks, proposed by Weston et al. \cite{weston2014memory}, which can explicitly store long-term memories and has easy access to reading memories. With such a mechanism, MRC models can have a deeper understanding of the context and the question by multiple turns of interaction. After being given the context as input, the memory mechanism stores information of the context in memory slots and then updates them dynamically. The process of answering involves finding the memory most relevant to the question and turning it into an answer representation as required. Although this method can overcome the shortcoming of insufficient memory, it is hard to train the network via back-propagation. To address this problem, Sukhbaatar et al. \cite{sukhbaatar2015end} introduce an end-to-end version of memory networks, in which explicit memory storage is embedded with continuous representations and the process of reading and updating memories is modeled by neural networks. This extension of memory networks can reduce supervision during training and is applicable to more tasks. 
\par The characteristic of updating memories with multiple hops in memory networks makes this method popular in MRC systems. Pan et al. \cite{pan2017memen} propose the MEMEN model, which stores question-aware context representations, context-aware question representations, and candidate answer representations in memory slots and updates them dynamically. Similarly, Yu et al. \cite{yu2018multi} use external memory slots to store question-aware context representations and update memories with bidirectional~GRUs.

\par The third method takes advantage of the recurrence feature of RNNs, using hidden states to store previous interaction information. Wang and Jiang \cite{wang2016machine} perform multiple interaction by using match-LSTM architecture recurrently. This model is originally proposed for textual entailment; when introduced to MRC, it can simulate the process of reading passages with question information. \mbox{First, Wang} and Jiang use the standard attention mechanism to obtain the attentive weight of each context token for the question. After calculating the dot product of question tokens and attentive weights, the model concatenates it with the context token and feeds it to a match-LSTM to get query-aware context representations. Similarly, this process is done in the reverse direction to fully encode contextual information. Finally, the outputs of match-LSTM in two directions are concatenated and fed to the answer prediction module. In addition, R-NET \cite{wang2017gated} and iterative alternating reader (\mbox{IA Reader}) \cite{sordoni2016iterative} also use RNNs to update query-aware context representations to perform multi-hop~interaction.

\par Some early work treated each context and query word equally when mining their correlations. However, the most important part should be given more attention for efficient context-query interaction. The gate mechanism, which can control the amount of mutual information between the context and the question, is a key component in multi-hop interaction.
\par For the gated-attention reader (GA Reader), Dhingra et al. \cite{dhingra2017gated} use the gate mechanism to decide how question information affects the focus on context words when updating context representations. The gate attention mechanism is performed by an element-wise multiplication between query embeddings and intermediate representations of context more than once.
\par In contrast to the GA Reader, context and question representations are updated in the iterative alternating attention mechanism \cite{sordoni2016iterative}. Question representations are updated with previous search states, while context representations are refined with previous reasoning information as well as currently updated queries. Then the gate mechanism, which is performed by a feed-forward network, is applied to determine the degree of matching between the context and the query. This mechanism is capable of extracting evidence from the context and the question alternately.
\par Previous models ignored the fact that context words have different importance when answering particular questions. To address this problem, Wang et al. \cite{wang2017gated} introduce the gate mechanism to filter out insignificant parts in the context and emphasize the ones most relevant to the question in their R-NET model. This model can be regarded as a variant of the attention-based recurrent networks. Compared to match-LSTM \cite{wang2016machine}, it introduces an additional gate mechanism based on current context representations and context-aware question representations. Moreover, as RNN-based models cannot deal well with long documents because of insufficient  memory, Wang et al. add self-attention to the context itself. This mechanism can dynamically refine context representations based on mutual information from the whole context and the question.
\par In conclusion, one-hop interaction may fail to comprehensively understand mutual question-context information. By contrast, multiple-hop interaction with the memory of previous contexts and questions is capable of deeply extracting correlations and aggregating evidence for answer~prediction.

\subsubsection{Answer Prediction}
This module is always the last of the MRC systems to give answers to questions according to the original context. The implementation of answer prediction is highly task-specific. Just as MRC tasks are categorized into cloze tests, multiple choice, span extraction, and free answering, there are four answer prediction methods: word predictor, option selector, span extractor, and answer generator. \mbox{In this part,} we will give an illustration in detail.

\noindent(1) Word Predictor
\par Cloze tests require filling in blanks with missing words or entities; one is asked to find out a word or an entity from the given context as the answer. In early work, such as the Attentive Reader \cite{hermann2015teaching}, \mbox{the combination} of the query-aware context and the question is reflected in the vocabulary space to search for the correct answer word. Chen et al. \cite{chen2016thorough} directly use query-aware context representations to match the candidate answer, simplifying the prediction process and improving the performance.
\par This method employs attentive context representations to select the correct answer word, but it cannot ensure that the answer is in the context, which does not satisfy the requirements of cloze tests. A related example is shown below (Kadle et al. \cite{kadlec2016text}).

\begin{table}[h]
	\centering
	\renewcommand{\arraystretch}{1.2}
	\begin{tabular}{l l}
		\textit{Context:} &\textit{ A UFO was observed above our city in January and again in March.}\\
		\textit{Question:} & \textit{An observer has spotted a UFO in \_\_\_\_\_.}\\
	\end{tabular}
\end{table}

\par  In the situation where both \textit{January} and \textit{March} can be the correct answer, methods used by Hermann et al. \cite{hermann2015teaching} and Chen et al. \cite{chen2016thorough}, which reflect attentive context representations into the whole vocabulary space, would give an answer similar to these two words, maybe \textit{February} because of features of distributed word representation pre-trained by Word2Vec.
\par To overcome the problem that the predicted answer may not be in the context, Kadlec et al. \cite{kadlec2016text} propose the attention sum reader (AS Reader) model, inspired by pointer networks. Pointer networks, introduced by Vinyals et al. \cite{vinyals2015pointer}, are adapted to tasks whose outputs can only be selected from inputs at first and can satisfy the requirements of cloze tests. In the AS Reader, Kadlec et al. do not compute the attentive representations; instead, they directly use attention weights to predict the answer. \mbox{The attention} results of the same word are added together, and the one with the maximum value is selected as the answer. This method is simple but quite efficient for cloze tests.

\noindent (2) Option Selector

\par To tackle the multiple-choice task, the model should select the correct answer from candidate options. The common way is to measure the similarity between attentive context representations and candidate answer representations and the most similar candidate is chosen as the correct answer.
\par Chaturvedi et al. \cite{chaturvedi2018cnn} use CNNs to encode question-option tuples and relevant context sentences. Then they measure the correlation between them by cosine similarity, and the most relevant option is selected as the answer. Zhu et al. \cite{zhu2018hierarchical} introduce the information of options to contribute to extracting the interaction between the context and the question. In the answer prediction module, they use a bilinear function to score each option according to the attentive information; the one with the highest score is the predicted answer. In the convolutional spatial attention model, Chen et al. \cite{chen2018convolutional} calculate similarities among question-aware candidate representations, context-aware representations, and self-attended question representations with dot product to fully extract correlations among context, question, and options. The various similarities are concatenated and then fed to CNNs with different kernel sizes. The outputs of the CNNs are regarded as feature vectors and fed to fully connected layers to calculate a score for each candidate. Finally, the correct answer is the one with highest score.

\noindent (3) Span Extractor
\par The span extraction task can be regarded as an extension of cloze tests, which requires extracting a subsequence from the context rather than a single word. As word predictor methods used in models for cloze tests can only extract one context token, they cannot be directly applied to the span extraction task. Also inspired by pointer networks \cite{vinyals2015pointer}, Wang and Jiang \cite{wang2016machine} propose two models, the sequence model and the boundary model to overcome the shortcomings of word prediction approaches. \mbox{Outputs of} the sequence model are positions where answer tokens appear in the original context. \mbox{The process} of answer prediction is similar to the decoding of sequence-to-sequence models, selecting tokens with the highest probability successively until it stops generating answer tokens. Answers obtained by such methods are treated as a sequence of tokens from the input context, which might not be a consecutive span and cannot be ensured to be a subsequence of original context. \mbox{The boundary} model can handle the problem, which only predicts the start and end positions of the answer. The boundary model is much simpler and shows better performance on SQuAD. It has been widely used in other MRC models as a preferred alternative for subsequence extraction.
\par Considering that there is more than one plausible answer span in the original context, but the boundary model might extract incorrect answers with local maxima, Xiong et al. \cite{xiong2016dynamic} proposed a dynamic pointing decoder to select an answer span by multiple iterations. This method uses LSTM to estimate the start and end positions based on representations corresponding to the last state answer prediction. To compute start and end score of context tokens, Xiong et al. propose highway maxout networks (HMN) with maxout networks \cite{goodfellow2013maxout} and highway networks \cite{srivastava2015highway}, which require different models according to various question types and context topics. 

\noindent (4) Answer Generator

\par With the appearance of free answering tasks, answers are no longer limited to a subsequence of the original context; instead, they need to be synthesized from the context and the question. Specifically, the expression of answers may differ from the evidence snippet in the given context, or answers may be from multiple pieces of evidence in different passages. Answer forms of the free answering task have the fewest limits, but this task has high requirements for the answer prediction module. To deal with the challenge, some generation approaches have been introduced to generate flexible~answers.
\par S-NET, proposed by Tan et al. \cite{tan2018s}, introduces the answer generation module to satisfy the requirement of free answering tasks, whose answers are not limited to the original context. It follows the ``extraction and then synthesis'' process. The extraction module is a variant of R-NET, \cite{wang2017gated} while the generation module is a sequence-to-sequence architecture. More specifically, for the encoder, bidirectional GRUs are used to produce context and question representations. In particular, start and end positions of evidence snippets predicted by the span extraction module are added to context representations as additional features. In terms of the decoder, the state of the GRUs is updated by the previous context word representations and attentive intermediate information. After applying the SoftMax function, the output of the decoder is the synthetic~answer. 
\par The generation module successfully makes up for the deficiency of the extraction module and generates more flexible answers. However, answers generated by existing approaches may suffer from syntax errors and faulty logic problems. Hence, generation and extraction methods are usually used together to provide complementary information. For example, in S-NET, the extraction module first labels the approximate boundary of the answer span, while the generation module generates answers not limited to the original context based on that. Generation approaches are not very common in existing MRC systems, as extraction methods already perform well enough in most cases.

\subsection{Additional Tricks}
While some typical deep-learning methods have been introduced here, there are some additional tricks, such as reinforcement learning, answer ranker, and sentence selector, that cannot be included in the general MRC architecture. However, these tricks also contribute to performance improvement, \mbox{so we} will give brief descriptions of them in the following sections.

\subsubsection{Reinforcement Learning}
\par As can be seen from the above introduction, most MRC models only apply maximum-likelihood estimation in the training process. However, there is a disconnect between optimization objectives and evaluation metrics. As a result, candidate answers that exactly match the ground truth or have word overlap with the ground truth but are not located at the labeled positions would be ignored by such models. In addition, when the answer span is too long or has fuzzy boundaries, models would also fail to extract the correct answer. MRC evaluation metrics, such as exact match (EM) and F1, are not differentiable, so some researchers introduce reinforcement learning to the training process. Xiong et al. \cite{xiong2017dcn+} and Hu et al. \cite{hu2018reinforced} use F1 score as the reward function and treat maximum-likelihood estimation and reinforcement learning as a multi-task learning problem. This method can consider both textual similarity and position information.

\par Reinforcement learning can also be used to determine whether to stop the interaction process. Multi-hop interaction methods, introduced above, have a pre-defined number of hops in the interaction. However, when people are answering a question, they stop reading if there is adequate evidence for the answer. The termination state is highly related to the complexity of the given context and question. Motivated by stopping interaction dynamically according to the context and question, Shen et al. \cite{shen2017reasonet} introduce a termination state to their ReasonNets. If the value of this state equals 1, the model stops the interaction and feeds the evidence to the answer prediction module to give the answer; otherwise ReasonNets continues the interaction by computing the similarity between the intermediate state and the input context and query. As the termination state is discrete and back-propagation cannot be used directly while training, reinforcement learning is applied to train the model by maximizing the instance-dependent expected reward. 

\par In a word, reinforcement learning can be regarded as an improved approach in MRC systems that is capable of not only reducing the gap between optimization objectives and evaluation metrics but also determining whether to stop reasoning dynamically. With reinforcement learning, the model can be trained  and refine better answers even if some states are discrete.

\subsubsection{Answer Ranker}
\par To verify whether the predicted answer is correct or not, some researchers introduce an answer ranker module. The common process of ranker is that it extracts some candidate answers, and the one with the highest score is the correct answer.
\par EpiReader \cite{trischler2016natural} combines pointer methods with the ranker. Trischler et al. extract answer candidates using an approach similar to the AS Reader \cite{kadlec2016text}, selecting some answer spans with the highest attention sum score. Then EpiReader feeds those candidates to the reasoner component, which inserts them to the question sequence at placeholder location and computes their probability to be the correct answers. The one with the highest probability is selected as the correct answer.
\par To extract candidates with variable lengths, \mbox{Yu et al.} \cite{yu2016end} propose two approaches. In the first one, they capture the part-of-speech (POS) patterns of answers in the training set and choose subsequences in the given passage that could match such patterns as candidates. The other approach enumerates all possible answer span within a fixed length from the context. After obtaining answer candidates, \mbox{Yu et al.} compute their similarity with question representations and choose the most similar one as the~answer.
\par With the ranker module, the accuracy of answer prediction can be improved somewhat. \mbox{These methods} have also inspired some researchers to detect unanswerable questions.

\subsubsection{Sentence Selector}
\par In practice, if the MRC model is given a long document, it takes a lot of times to understand the full context to answer questions. However, finding the sentences that are most relevant to the questions in advance is a possible way to accelerate the training process. With this motivation, Min et al. \cite{min2018efficient} propose a sentence selector to find the minimal set of sentences needed to answer a question. \mbox{The architecture} of the sentence selector is sequence-to-sequence, which contains an encoder to compute sentence and question encodings and a decoder to calculate a score for each sentence by measuring the similarity between the sentence and the question. If the score is higher than a pre-defined threshold, the sentence is selected to be fed to the MRC systems. In this way, the number of selected sentences is dynamic according to different questions.
\par An MRC system with a sentence selector is capable of reducing training and inference time with equivalent or better performance compared to a system without a sentence selector.
\section{Datasets and Evaluation Metrics} \label{s4}
The release of large-scale MRC datasets makes it possible to train deep neural models while evaluation metrics can evaluate the performance of MRC models, both of which play significant role in the MRC field. In this section, we describe representative datasets according to different tasks,\mbox{ then introduce} evaluation metrics.

\subsection{Datasets}
Datasets are among the driving factors accelerating the development of the MRC field, some of which, such as CNN \& Daily Mail \cite{hermann2015teaching}, SQuAD \cite{rajpurkar2016squad} and MS MARCO \cite{nguyen2016ms}, can be regarded as milestones of MRC, spurring the emergence of up-to-date techniques. In this part, we introduce several representative datasets of each MRC task, highlighting how to construct large-scale datasets according to task requirements, and how to reduce lexical overlap between questions and context.

\subsubsection{Cloze Tests Datasets}
\begin{itemize}[itemindent=2em,listparindent=2em,leftmargin=0pt,itemsep=10pt,parsep=10pt]
	\item[-] CNN \& Daily Mail
	\par This dataset, built by Hermann et al. \cite{hermann2015teaching}, is one of the most representative cloze-style MRC datasets. CNN \& Daily Mail, consisting of 93,000 articles from the CNN and 220,000 articles from the Daily Mail, is indeed large-scale and makes it possible to use deep-learning approaches in MRC. Considering that bullet points are abstractive and have little sentence overlap with documents, Hermann et al. replace one entity at a time with a placeholder in bullet points and evaluate the machine reading system by asking the machine to read the documents and then predict which entity the placeholder in the bullet points referred to. As questions are not posed directly from documents, this task is challenging, and some information extraction methods fail to deal with it. This methodology of creating MRC datasets has enlightened a lot of other research\cite{suster2018clicr,onishi2016did,saha2018duorc}. To avoid the situation where questions can be answered by knowledge outside of the documents, all entities in documents are anonymized by random markers.
	
	\item[-] CBT
	\par Hill et al.\cite{hill2015goldilocks} design a cloze-style MRC dataset, the Children's Book Test (CBT), from another perspective. They collect 108 children's books and form each sample with 21 consecutive sentences from chapters in those books. To generate questions, a word from the 21st sentence is removed, and the other 20 sentences act as context. Nine incorrect words, whose type is the same as the answer, are selected at random from the context as candidate answers. There are some differences between the CNN \& Daily Mail and CBT dataset. First, unlike the CNN \& Daily Mail, entities in the CBT dataset are not anonymized, so models can use background knowledge from wider contexts. Second, missing items in the CNN \& Daily Mail are limited to named entities, but in the CBT there are four distinct types: named entities, nouns, verbs, and prepositions. Third, the CBT provides candidate answers, which simplifies the task in a way. Overall, with the appearance of the CBT, context, which plays a significant role in human comprehension, has gained much more attention. Considering that more data can significantly improve the performance of neural network models, Bajgar et al. \cite{bajgar2016embracing} introduce the BookTest, which enlarges the CBT dataset 60 times and enables training of larger models.
	\item[-] LAMBADA
	\par To account for the meaning of the wider context, Paperno et al. \cite{paperno2016lambada} propose the language modeling broadened to account for discourse aspects (LAMBADA) dataset. Similar to the CBT, LAMBADA also uses books as the source, and the task is word prediction. However, the word that needs to be predicted in LAMBADA is the last word in the target sentence, while in CBT any word in the target sentence may be targeted. Moreover, Paperno et al. find that some samples in the CBT can be guessed with the target sentence alone rather than in a wider context. To overcome this shortcoming, there is a constraint in LAMBADA that makes it difficult to predict the target word correctly only with the target sentence. That is to say, compared with CBT, LAMBADA requires more understanding of the wider context.
	\item[-] Who-did-What
	\par To better evaluate the understanding of natural language, researchers try to avoid sentence overlap between questions and documents when constructing MRC datasets. Onishi et al. \cite{onishi2016did} provide new insight into how to reduce the syntactic similarity. In the ``Who-did-What'' dataset, each sample is formed from two independent articles; one serves as the context and questions are generated from the other. This approach can be used by other corpora, in which articles do not have summary points, unlike CNN \& Daily Mail. There is another feature of Who-did-What, as shown in the name; the dataset only pays attention to the personal name entity, which may be its limitation.
	
	\item[-] CLOTH
	\par In contrast to the above automatically generated datasets, cloze test by teachers (CLOTH) \cite{xie2018large} is human-created, collected from English exams for Chinese students. Questions in CLOTH are well-designed by middle-school and high-school teachers to examine students' language proficiency, including vocabulary, reasoning, and grammar. There are fewer purposeless or trivial questions in CLOTH, so that it requires a deep understanding of language.
	\item[-] CliCR
	\par To address the problem of scarce datasets for specific domains, Suster et al.\cite{suster2018clicr} build a large-scale cloze-style dataset based on clinical case reports for healthcare and medicine. Similar to the CNN \& Daily Mail, summary points of case reports are used to create queries by blanking out medical entities. CliCR promotes MRC to practical applications such as clinical decision-making.
\end{itemize}

\subsubsection{Multiple-Choice Datasets}
\begin{itemize}[itemindent=2em,listparindent=2em,leftmargin=0pt,itemsep=10pt,parsep=10pt]
	\item[-] MCTest
	\par MCTest, proposed by Richardson et al. \cite{richardson2013mctest}, is a multiple-choice MRC dataset at an early stage. \mbox{It consists} of 500 fictional stories, and for each story there are four questions with four candidate answers. Choosing fictional stories avoids introducing external knowledge, and questions can be answered according to the stories themselves. This idea of using a story-based corpus has inspired other datasets, such as CBT \cite{hill2015goldilocks} and LAMBADA \cite{paperno2016lambada}. Although the appearance of MCTest has encouraged research on MRC, its size is too small, so it is not suitable for some data-hungry techniques.
	\item[-] RACE
	\par Like the CLOTH dataset \cite{xie2018large}, RACE \cite{lai2017race} is also collected from English exams for middle-school and high-school Chinese students. This corpus allows a greater variety of types of passages. \mbox{In contrast} to one fixed style for the whole dataset, such as news for CNN \& Daily Mail \cite{hermann2015teaching} and NewsQA \cite{trischler2017newsqa}, and fictional stories for CBT \cite{hill2015goldilocks} and MCTest \cite{richardson2013mctest}, almost all kinds of passages can be found in RACE. As a multiple-choice task, RACE asks for more reasoning, because questions and answers are human-generated and simple methods based on information retrieval or word co-occurrence may not perform well. In addition, compared to MCTest \cite{richardson2013mctest}, RACE, which contains about 28,000 passages and 100,000 questions, is large-scale and supports the training of deep-learning models. All the aforementioned features illustrate that RACE is well-designed and full of challenges.
\end{itemize}
\newpage

\subsubsection{Span Extraction Datasets}
\begin{itemize}[itemindent=2em,listparindent=2em,leftmargin=0pt,itemsep=10pt,parsep=10pt]
	\item[-] SQuAD
	\par The Stanford Question-Answering Dataset (SQuAD), proposed by Rajpurkar et al.\cite{rajpurkar2016squad} of Stanford University, can be regarded as a milestone of MRC. With the release of SQuAD dataset, an MRC competition based on it has drawn the attention of both academia and industry, which in turn has stimulated the development of various advanced MRC techniques.  Collecting 536 articles from Wikipedia, Rajpurkar et al. ask crowd-workers to pose more than 100,000 questions and select a span of arbitrary length from the given article to answer the question. SQuAD is not only large but also high quality. In contrast to prior datasets, SQuAD defines a new kind of MRC task, which does not provide answer choices, but needs a span of text as the answer rather than a word or an entity. 
	\item[-] NewsQA
	\par NewsQA \cite{trischler2017newsqa} is another span extraction dataset similar to SQuAD, in which questions are also human-generated and answers are spans of text from corresponding articles. The obvious difference between NewsQA and SQuAD is the source of articles. In NewsQA, articles are collected from CNN, while the SQuAD is based on Wikipedia. It is worth mentioning that some questions in NewsQA have no answer according to the given context. The addition of unanswerable questions makes it closer to reality and inspires Rajpurkar et al. \cite{rajpurkar2018know} to update SQuAD to version 2.0. In terms of unanswerable questions, we give a detailed introduction in Section \ref{noanswer}.
	\item[-] TriviaQA
	\par The construction process of TriviaQA \cite{joshi2017triviaqa} distinguishes it from previous datasets. In prior work, crowd-workers are given articles and pose questions closely related to those articles. However, this process results in the dependence of questions and evidence to answer them. Furthermore, in human understanding, people often ask a question and then find useful resources to answer it. To overcome this shortcoming, Joshi et al. gather question-answer pairs from trivia and quiz-league websites. \mbox{Then they} search for evidence to answer questions from webpages and Wikipedia. Finally, they build more than 650,00 question-answer-evidence triples for the MRC task. This novel construction process makes TriviaQA a challenging testbed with considerable syntactic variability between questions and~contexts.
	\item[-] DuoRC
	\par Saha et al. \cite{saha2018duorc} also try to reduce lexical overlap between questions and contexts in DuoRC. \mbox{As in} Who-did-What \cite{onishi2016did}, questions and answers in DuoRC are created from two different versions of documents corresponding to the same movie, one from Wikipedia and one from IMDb. \mbox{Asking questions} and labeling answers are done by different group of crowd-workers. The distinction between the two versions of movie plots requires more understanding and reasoning. Moreover, there are unanswerable questions in DuoRC.
\end{itemize}

\subsubsection{Free Answering Datasets}
\begin{itemize}[itemindent=2em,listparindent=2em,leftmargin=0pt,itemsep=10pt,parsep=10pt]
	\item[-] bAbI
	\par bAbI, proposed by Weston et al. \cite{weston2015towards}, is a well-known synthetic MRC dataset. It consists of 20~tasks, generated with a simulation of a classic text adventure game. Each task is independent from the others and tests one aspect of text understanding, such as recognizing two or three argument relations and using basic deduction and induction. Weston et al. think that dealing with all these tasks is a prerequisite to full language understanding. Answers are limited to a single word or a list of words and may not be directly found from the original context. The release of the bAbI dataset promotes the development of several promising algorithms, but as all data in bAbI is synthetic, it is a little far from the real world. 
	\item[-] MS MARCO
	\par MS MARCO \cite{nguyen2016ms} can be viewed as another milestone of MRC after SQuAD \cite{rajpurkar2016squad}. To overcome the weaknesses of previous datasets, it has four predominant features. First, all questions are collected from real user queries. Second, for each question, 10 related documents are searched with the Bing search engine to serve as the context. Third, labeled answers to those questions are generated by humans so that they are not restricted to spans of the context and more reasoning and summarization are required. Fourth, there are multiple answers to each question and sometimes they even conflict, which makes it more challenging for the machine to select the correct answer. MS MARCO makes the MRC dataset closer to the real world.
	\item[-] SearchQA
	\par The work of SearchQA \cite{dunn2017searchqa} is just like TriviaQA \cite{joshi2017triviaqa}; both follow the general pipeline of question answering. To construct SearchQA, Dunn et al. collect question-answer pairs from the J!Archive and then search for snippets related to questions from Google. However, the major difference between SearchQA and TriviaQA is that in TriviaQA there is one document with evidence for each question-answer pair, while in SearchQA each pair has 49.6 related snippets on average.
	\item[-] NarrativeQA
	\par Seeing the limitation in most previous datasets of needing evidence to answer questions from single sentences of original context, Kovcisky et al. \cite{kovcisky2018narrativeqa} design the NarrativeQA. Based on book stories and movie scripts, they search related summaries from Wikipedia and ask co-workers to generate question-answer pairs according to those summaries. What makes NarrativeQA special is that answering questions requires understanding the whole narrative, rather than superficial matching information.
	\item[-] DuReader
	\par Similar to MS MARCO \cite{nguyen2016ms}, DuReader, released by He et al. \cite{he2018dureader},  is another large-scale MRC dataset from real-world application. Questions and documents in DuReader are collected from Baidu Search (search engine) and Baidu Zhidao (question-answering community). Answers are human-generated instead of spans of original context. What makes DuReader different is that it provides new question types such as yes/no and opinion. Compared to factoid questions, these questions sometimes require summaries over multiple parts of documents, which opens an opportunity for the research community. 
\end{itemize}

\subsection{Evaluation Metrics}
For different MRC tasks, there are various evaluation metrics. To evaluate cloze tests and multiple-choice tasks, the most common metric is accuracy. In terms of span extraction, exact match (EM), a variant of accuracy, and F1 score are computed to measure model performance. Considering that answers for free answering tasks are not limited to the original context, ROUGE-L and BLEU are widely used. In the following part, we will give detailed descriptions of these evaluation metrics.
\begin{itemize}[itemindent=2em,listparindent=2em,leftmargin=0pt,itemsep=10pt,parsep=10pt]
	\item[-] Accuracy
	\par Accuracy with respect to ground-truth answers is usually applied to evaluate cloze tests and multiple-choice tasks. When given a question set $Q=\{Q_{1},Q_{2},\cdots Q_{m}\}$ with $m$ questions, if the model correctly predicts answers for $n$ questions, then accuracy is calculated as follows:
	
	\begin{equation}
	\text{Accuracy} = \frac{n}{m}.
	\end{equation} 
	
	\par Exact match is a variant of accuracy that evaluates whether a predicted answer span matches the ground-truth sequence exactly or not. If the predicted answer is equal to the gold answer, the EM value will be 1, otherwise 0. It can also be calculated by the above equation.
	\item[-] F1 Score
	\par F1 score is a common metric in classification tasks. In terms of MRC, both candidate and reference answers are treated as bags of tokens and true positive (TP), false positive (FP), true negative (TN), and false negative (FN) are denoted as shown in Table \ref{f1}.
	
	\begin{table}[h]
		\centering
		\renewcommand{\arraystretch}{1.2}
		\caption{\upshape The definition of true positive (TP), true negative (TN), false positive (FP), false negative (FN).}
		\label{f1}
		\begin{tabular}{ccc}
			\toprule
			& \textbf{Tokens in Reference} & \textbf{Tokens Not in Reference}\\
			\midrule
			tokens in candidate & TP & FP\\ 
			tokens not in candidate & FN & TN \\
			\bottomrule
		\end{tabular}
	\end{table}
	
	\par Then precision and recall are computed as below:
	
	\begin{equation}
	\text{precision}=\frac{\text{TP}}{\text{TP}+\text{FP}}\;,
	\end{equation}
	
	\begin{equation}
	\text{recall}=\frac{\text{TP}}{\text{TP}+\text{FN}}.
	\end{equation}
	
	F1 score, also known as balanced F score, is the harmonic average of precision and recall: 
	
	\begin{equation}
	\text{F1}=\frac{2 \times P \times R}{P+R}\;,
	\end{equation}
	where $P$ denotes precision while $R$ is recall.
	\vspace{-10pt}
	\par Compared to EM, this metric loosely measures the average overlap between the prediction and the ground-truth answer.

	\item[-] ROUGE-L
	\par ROUGE (Recall-Oriented Understudy for Gisting Evaluation) is an evaluation metric initially developed for automatic summarization, proposed by Lin and Chin-Yew \cite{lin2004rouge}. It evaluates the quality of a summary by counting the amount of overlap between model-generated and ground-truth summaries. There are various ROUGE measures, such as ROUGE-N, ROUGE-L, ROUGE-W, and ROUGE-S, for~different evaluation requirements, among which ROUGE-L is widely used in MRC tasks with the use of free answering. Unlike other metrics, such as EM or accuracy, ROUGE-L is more flexible, mainly measuring similarities between gold answers and predicted answers. The ``L'' in ROUGE-L denotes longest common subsequence (LCS) and ROUGE-L can be computed as follows:
	
	\begin{equation}
	R_{\text{lcs}}=\frac{LCS(X,Y)}{m},
	\end{equation}
	
	\begin{equation}
	P_{\text{lcs}}=\frac{LCS(X,Y)}{n},
	\end{equation}
	
	\begin{equation}
	F_{\text{lcs}}=\frac{(1+\beta)^{2}R_{\text{lcs}}P_{\text{lcs}}}{R_{\text{lcs}}+\beta^{2}P_{\text{lcs}}},
	\end{equation}
	where $ X $ is ground-truth answer with $ m $ tokens, $ Y $ is model-generated answer with $ n $ tokens, $\text{LCS}(X,Y)$  denotes the length of the longest common subsequence of $X$ and $Y$, and $\beta$ is a parameter to control the importance of precision $P_{\text{lcs}}$ and recall $R_{\text{lcs}}$.
	
	\vspace{-10pt}
	Using ROUGE-L to evaluate the performance of an MRC model does not require predicted answers to be consecutive subsequences of ground truth, whereas more token overlap contributes to higher ROUGE-L scores. However, the length of candidate answers influences the value of~ROUGE-L.
	
	\item[-] BLEU
	
	BLEU (Bilingual Evaluation Understudy), proposed by Papineni et al. \cite{papineni2002bleu}, is widely used to evaluate translation performance. When adapted to MRC tasks, BLEU score measures the similarity between predicted answers and ground truth. The cornerstone of this metric is precision measure, which is calculated as follows:
	
	\begin{equation}
	P_{n}(C,R) = \frac{\sum_{i}\sum_{k}\text{min}(h_{k}(c_{i}),\text{max}(h_{k}(r_{i})))}{\sum_{i}\sum_{k}h_{k}(c_{i})},
	\end{equation}
	where $h_{k}(c_{i})$ counts the number of $k$-th n-gram appearing in candidate answer $c_{i}$; in a similar way, $h_{k}(r_{i})$ denotes the occurrence number of that n-gram in gold answer $r_{i}$.
	
	\vspace{-10pt}
	As the value of $P_{n}(C,R)$ is higher when the answer span is shorter, such precision cannot measure the similarity well by itself. The penalty factor, BP, is introduced to alleviate that, which is computed~as:
	
	\begin{equation}
	\text{BP}=
	\begin{cases}
	1,l_{c}>l_{r}\\
	e^{1-\frac{l_{r}}{l_{c}}},l_{c}\leq l_{r}.
	\end{cases}
	\end{equation}
	
	Finally, the BLEU score is computed as follows:
	
	\begin{equation}
	\text{BLEU}=\text{BP} \cdot \text{exp}(\sum_{n=1}^N w_{n}\log P_{n}),
	\end{equation}
	where $N$ means n-grams up to length $N$ and $w_{n}$ equals $1/N$. The BLEU score is the weighted average of each n-gram and the maximum of $N$ is 4, namely BLEU-4.
	
\end{itemize}
\vspace{-8pt}
\par BLEU score can not only evaluate the similarity between candidate answers and ground-truth answers but also test the readability of candidates.

\section{New Trends} \label{s5}
Since about 2017, the research community has started to focus on developing neural models for some more complicated problems or tasks related to MRC, such as how to leverage knowledge bases and how to deal with unanswerable questions, with the belief that they are more significant for practical applications and that deep-learning techniques can address them more effectively, though some of them may have been involved in traditional expert or QA systems. We regard them as new trends within the context of neural MRC and discuss them in the following sections.
\subsection{Knowledge-Based Machine Reading Comprehension}

MRC requires answering questions with knowledge implicit in the given context. Datasets such as the MCTest choose passages from specific corpora (fiction stories, children's books, etc.) to avoid introducing external knowledge. However, human-generated questions are usually too simple when compared to questions in real-world application. In the process of human reading comprehension, we~may use common sense when we cannot answer a question simply by knowing about the context. External knowledge is so significant that is believed to be the biggest gap between MRC and human reading comprehension. As a result, interest in introducing world knowledge to MRC has surged in the research community, and knowledge-based machine reading comprehension (KBMRC) has come into being. KBMRC differs from MRC mainly in terms of inputs. In MRC, the inputs are sequences of context and questions. However, besides that, additional related knowledge extracted from knowledge bases is necessary in KBMRC. 
\par KBMRC can be regarded as augmented MRC with external knowledge $K$ and it can be formulated as shown in Table \ref{d_knowledge}:

\vspace{12pt}

\begin{table}[h]
	\renewcommand{\arraystretch}{1.5}
	\centering
	\caption{Definition of knowledge-based machine reading comprehension.}
	\label{d_knowledge}
	\begin{tabular}{p{12cm}}
		\noalign{\hrule height 1pt}
		\bfseries{Knowledge-Based Machine Reading Comprehension}\\
		\hline
		\rowcolor{tabcolor}Given the context $C$, question $Q$ and external knowledge $K$, the task requires predicting the correct answer $A$ by learning the function $\mathcal{F}$ such that $A=\mathcal{F}(C,Q,K)$.\\
		\noalign{\hrule height 1pt}
	\end{tabular}
\end{table}

\par There are some KBMRC datasets in which world knowledge is needed to answer some questions. MCScripts \cite{ostermann2018mcscript} is a dataset about human daily activities, such as eating in a restaurant and taking a bus, where answering some questions requires common sense knowledge beyond the given context. As shown in Table \ref{KBMRC}, the answer to \textit{Why are trees important?} cannot be found in the given context. However, it is common sense knowledge to us that trees are important because they \textit{create oxygen} and absorb carbon dioxide via photosynthesis not because \textit{they are green}.

\begin{table}[h]
	\centering
	\renewcommand{\arraystretch}{1.2}
	\caption{\upshape Some Examples in KBMRC}
	\label{KBMRC}
	\begin{tabular}{ll}
		\hline
		\rowcolor{tabcolor}\multicolumn{2}{l}{\bfseries{MCScripts}}\\
		\hline
		Context: & \makecell[l]{Before you plant a tree, you must contact the utility company.\\ They will come to your property and mark out utility lines.\\ Without doing this, you may dig down and hit a line, which\\ can be lethal! Once you know where to dig, select what type\\ of tree you want. Take things into consideration such as how \\much sun it gets, what zone you are in, and how quickly you\\ want it to grow. Dig a hole large enough for the tree and roots.\\ Place the tree in the hole and then fill the hole back up with dirt.}\\
		\hline
		Question: & Why are trees important?\\
		Candidate Answers: & \color{blue}A. create $O_{2}$ \ \ \ \color{black}B. because they are green\\
		\hline
	\end{tabular}
\end{table} 

\par The key challenges in KBMRC are as follows:
\begin{itemize}[itemindent=2em,listparindent=2em,leftmargin=0pt,itemsep=10pt,parsep=10pt]
	\item[-] Relevant External Knowledge Retrieval
	\par There are various kinds of knowledge stored in knowledge bases, and entities may be misleading sometimes because of polysemy, e.g., ``apple'' can refer to a fruit or a corporation. Extracting knowledge closely related to the context and the question determines the performance of knowledge-based answer~prediction. 
	\item[-] External Knowledge Integration
	\par In contrast to text in the context and questions, knowledge in an external knowledge base has its own unique structure. How to encode such knowledge and integrate it with representations of the context and questions remains an ongoing research challenge.
\end{itemize}

Some researchers have tried to address the above challenges in KBMRC. To make the model take advantage of the external knowledge, Long et al. \cite{long2017world} propose a new task, rare entity prediction, which requires predicting missing name entities and is similar to cloze tests. However, name entities removed from the context cannot be predicted correctly based only on the original context. \mbox{This task} provides additional entity descriptions extracted from knowledge bases such as Freebase as external knowledge to help with entity prediction. While incorporating external knowledge, \mbox{Yang and Mitchell \cite{yang2017leveraging}} consider the relevance between knowledge and context to avoid irrelevant external knowledge misleading the answer prediction. They design the attention mechanism with a sentinel to determine whether to incorporate external knowledge or not and which knowledge should be adopted. Mihaylov and Frank \cite{mihaylov2018knowledgeable} and Sun et al. \cite{sun2018knowledge} use key-value memory networks~\cite{miller2016key} to determine relevant external knowledge. All possible related knowledge is first selected from a knowledge base and stored in memory slots as key-value pairs. Then keys are used to match with the query, while corresponding values are summed together according to different weights to generate relevant knowledge representations. Wang and Jiang \cite{wang2019explicit} propose a data enrichment method with semantic relations in WordNet, a lexical database for English. For each word in the context and question, they try to find the positions of passage words that have a direct or indirect semantic relationship with it to. This position information is regarded as external knowledge and fed to the MRC model to assist answer prediction.
\par In conclusion, KBMRC breaks through the limitation that the scope of knowledge required to answer questions is restricted to the given context. Hence, this task, benefit from external world knowledge, can close the gap between machine comprehension and human understanding to some extent. However, the performance of KBMRC systems is highly related to the quality of the knowledge base. Disambiguation is required when extracting related external knowledge from automated or semi-automated knowledge bases, as entities with the same name or alias could mislead the models. Moreover, knowledge stored in knowledge bases is usually sparse. If related knowledge cannot be found directly, incorporating external knowledge calls for further inference.

\subsection{Machine Reading Comprehension with Unanswerable Questions} \label{noanswer}
There is a latent hypothesis behind MRC tasks that correct answers always exist in the given context. However, this does not conform with real-world application. The range of knowledge covered in a passage is limited; thus, some questions inevitably have no answers according to the given context. A mature MRC system should distinguish those unanswerable questions. 
\par MRC with unanswerable questions considers questions that cannot be answered based on the given context in a process consisting of two subtasks: answerability detection and reading comprehension. The definition of this new task is shown in Table \ref{d_un}:

\vspace{12pt}
\begin{table}[h]
	\renewcommand{\arraystretch}{1.5}
	\centering
	\caption{Definition of machine reading comprehension with unanswerable questions.}
	\label{d_un}
	\begin{tabular}{p{12cm}}
		\noalign{\hrule height 1pt}
		\bfseries{Machine Reading Comprehension with Unanswerable Questions}\\
		\hline
		\rowcolor{tabcolor}Given the context $C$ and question $Q$, the machine first determines whether $Q$ can be answered or not based on the given context $C$. If the question is impossible to be answered, the model marks it as unanswerable and abstain from answering, otherwise predicts the correct answer $A$ by learning the function $\mathcal{F}$ such that $A=\mathcal{F}(C,Q)$.\\
		\noalign{\hrule height 1pt}
	\end{tabular}
\end{table}
\vspace{12pt}

\par SQuAD 2.0 \cite{rajpurkar2018know} is a representative MRC dataset with unanswerable questions. Based on the previous version released in 2016, SQuAD 2.0 has more than 50,000 unanswerable questions created by crowd-workers. Those questions, impossible to answer based on the context alone, are challenging, as they are relevant to the given context and there are plausible answer spans matches the question requirements. To perform well on SQuAD 2.0, a model should not only gives correct answers to answerable questions, but also detect which questions have no answers. An example of an unanswerable question in SQuAD 2.0 is presented in Table \ref{unanswerable}. In the context, the keywords \textit{1937 treaty} exist and \textit{Bald Eagle Protection Act} is the name of the treaty in 1940, not 1937, which is very~puzzling.

\begin{table}[h]
	\centering
	\renewcommand{\arraystretch}{1.2}
	\caption{\upshape Unanswerable question example in SQuAD 2.0}
	\label{unanswerable}
	\scalebox{.95}[.95]{\begin{tabular}{ll}
			\noalign{\hrule height 1pt}
			\rowcolor{tabcolor}\multicolumn{2}{l}{\bfseries{SQuAD 2.0}}\\
			\hline
			Context: & …… Other legislation followed, including the Migratory Bird\\
			& Conservation Act of 1929, a \color{blue}1937 treaty \color{black} prohibiting the hunting\\ 
			&of right and gray whales, and the \color{red}Bald Eagle Protection Act of\\
			& \color{red}1940\color{black}. These later laws had a low cost to society---the species\\
			& were relatively rare---and little opposition was raised. \\
			\hline
			Question: & What was the name of the 1937 treaty\\
			Plausible Answer: & Bald Eagle Protection Act\\
			\noalign{\hrule height 1pt}
	\end{tabular}}
\end{table} 

\par With unanswerable questions, there are two other challenges in this new task, compared to MRC:
\begin{itemize}[itemindent=2em,listparindent=2em,leftmargin=0pt,itemsep=10pt,parsep=10pt]
	\newpage
	\item[-] Unanswerable Question Detection
	\par The model should know what it does not know. After comprehending the question and reasoning through the passage, the MRC models should judge which questions are impossible to answer based on the given context and mark them as unanswerable.
	\item[-] Plausible Answer Discrimination
	\par To avoid the impact of fake answers, such as the example presented in Table \ref{unanswerable}, the MRC model must verify the predicted answers and tell plausible answers from correct ones.
\end{itemize}
\vspace{-8pt}
\par For the above two challenges, methods applied to tackle the problems in MRC with unanswerable questions can be categorized into two sorts:
\par To indicate no-answer cases, one approach employs a shared-normalization operation between a no-answer score and an answer span score. Levy et al. \cite{levy2017zero} add an extra trainable bias to the confidence score of start and end position and apply SoftMax to the new score to obtain the probability distributions of no answer. If this probability is higher than that of the best span, it means the question is unanswerable, otherwise it outputs the answer span. In addition, they propose another method that sets a global confidence threshold; if the predicted answer confidence is below the threshold, the model labels the question as unanswerable. Although this approach can detect unanswerable questions, it~cannot guarantee that predicted answers are correct. The other methods introduce a no-answer option by padding. Tan et al. \cite{tan2018know} add a padding position for the original passage to determine whether the question is answerable. When the model predicts that position, it refuses to give an answer.
\par Researchers also pay attention to the legitimacy of answers and have introduced answer verification to discriminate plausible answers. For unanswerable question detection, Hu et al. \cite{hu2019read+} propose two types of auxiliary loss, independent span loss to predict plausible answers regardless of the answerability of the question, and independent no-answer loss, which alleviates the conflict between plausible answer extraction and no-answer detection tasks . In terms of answer verification, they introduce three methods. The first, sequential architecture, treats the question, answer, context sentence containing candidate answers as a whole sequence, and inputs that to the fine-tuned transformer model to predict the no-answer probability. The second is interactive architecture, which calculates the correlation between question-and-answer sentence in the context to classify whether the question is answerable. The third method integrates the above two approaches by concatenating the outputs of two models as joint representations, and this hybrid architecture can yield better performance. 
\par In contrast to the above pipeline structure, Sun et al. \cite{sun2018u} use multi-task learning to jointly train answer prediction, no-answer detection, and answer validation. What distinguishes their work is a combined universal node encoding passage and question information, which is then integrated with question representations and answer position aware passage representations. After being passed through the linear classification layer, fused representations can be used to determine whether the questions are answerable.
\par As the Chinese saying goes, \textit{To know what it is that you know, and to know what it is that you do not know, that is wisdom.} The detection of unanswerable questions requires deep understanding of text and requires more robust MRC models, making MRC much closer to real-world application.

\subsection{Multi-Passage Machine Reading Comprehension}
In MRC tasks, the relevant passages are pre-identified, which contradicts the question-answering process of humans. People usually ask a question first and then search for possibly related passages, where they find evidence to give the answer. To overcome this shortcoming, Chen et al. \cite{chen2017reading} extend MRC to machine reading at scale, more widely called multi-passage machine reading comprehension, which does not give one relevant passage for each question, unlike the traditional task. This extension can be applied to tackle open-domain question-answering tasks based on a large corpus of unstructured text. With its appearance, some multi-passage MRC task-specific datasets have been released, such as MS MARCO \cite{nguyen2016ms}, TriviaQA \cite{joshi2017triviaqa}, SearchQA \cite{dunn2017searchqa}, DuReader \cite{he2018dureader}, and QUASAR \cite{dhingra2017quasar}.
\par In contrast to MRC, in a multi-passage MRC setting, the given context $C$ is not a single passage, but a collection of documents $\mathbb{D}$. The definition of multi-passage MRC tasks changes as shown in Table \ref{d_multi}:
\vspace{12pt}
\begin{table}[h]
	\renewcommand{\arraystretch}{1.5}
	\centering
	\caption{Definition of multi-passage machine reading comprehension.}
	\label{d_multi}
	\begin{tabular}{p{12cm}}
		\noalign{\hrule height 1pt}
		\bfseries{Multi-Passage Machine Reading Comprehension}\\
		\hline
		\rowcolor{tabcolor}Given a collection of $m$ documents $\mathbb{D}=\{D_{1},D_{2},\cdots,D_{m}\}$ and the question $Q$, the multi-passage MRC task requires giving the correct answer $A$ to question $Q$ according to documents $\mathbb{D}$ by learning the function $\mathcal{F}$ such that $A=\mathcal{F}(\mathbb{D},Q)$.\\
		\noalign{\hrule height 1pt}
	\end{tabular}
\end{table}
\vspace{12pt}

\par Compared to MRC tasks, multi-passage MRC is far more challenging. For instance, although the DrQA model \cite{chen2017reading} achieves exact match accuracy of 69.5 on SQuAD, when applied to an open-domain setting (using the whole Wikipedia corpus to answer the question), its performance drops dramatically. The unique features of multi-passage MRC listed below are the main reasons for the degradation:

\begin{itemize}[itemindent=2em,listparindent=2em,leftmargin=0pt,itemsep=10pt,parsep=10pt]
	\item[-] Massive Document Corpus 
	\par This is the most prominent feature of multi-passage MRC, which makes it distinct from MRC, with one related passage. Under this circumstance, whether a model can retrieve the most relevant documents from the corpus quickly and correctly decides the final performance of reading comprehension. 
	\item[-] Noisy Document Retrieval
	\par Multi-passage MRC can be regarded as a distantly supervised open-domain question-answering task that may suffer from noise issues. Sometimes the model may retrieve a noisy document that contains the correct answer span, but it is not related to the question. This noise will mislead the understanding of the context.
	\item[-] No Answer
	\par When the retrieval component does not perform well, there will be no answers in the document. If the answer extraction module ignores that, it outputs an answer even if it is incorrect, which will lead to performance degradation.
	\item[-] Multiple Answers
	\par In the open-domain setting, it is common to find multiple answers for a single question. \mbox{For example}, when asking \textit{Who is the president of the United States}, both \textit{Obama} and \textit{Trump} are possible answers, but which is the correct answer requires reasoning based on the context.
	\item[-] Evidence Aggregation
	\par In terms of some complicated questions, snippets of evidence can appear in different parts of one document or even in different documents. To answer such questions correctly, a multi-passage MRC model needs to aggregate all the evidence. More documents mean more information, which contributes to more complete answers.
\end{itemize}
\vspace{-11pt}
\par To address multi-passage MRC problems, one method follows the pipeline of ``retrieve then read.'' More specifically, the retrieval component first returns several relevant documents, which are then proposed by the reader to give the answer. DrQA, introduced by Chen et al. \cite{chen2017reading}, is a typical pipeline-based multi-passage MRC model. In the retrieval component, they use TF-IDF to select  five relevant Wikipedia articles for each question in SQuAD to narrow the search space. For the reader module, they improve the model proposed in 2016 \cite{chen2016thorough} with rich word representations and a pointer module to predict the begin and end positions of answer spans. To make scores of candidate spans throughout different passages comparable, Chen et al. use unnormalized exponential and argmax functions to choose the best answer. In this approach, retrieval and reading are performed separately, but errors made in the retrieval stage are easily propagated to the next reading component, which leads to performance degradation. 
\par To alleviate error propagation caused by poor document retrieval, one way is to introduce \mbox{the ranker} component, and the other is to jointly train the retrieval and reading processes. 
\par In terms of the ranker component, which re-ranks the documents retrieved by the search engine, Htut et al. \cite{htut2018training} introduce two rankers, InferSent and Relation-Networks Rankers. The first uses a feed-forward network to measure the general semantic similarity between the context and question, while the second applies relation-networks to capture local interactions between context words and question words. Inspired by Learning to Rank research, Lee et al. \cite{lee2018ranking} propose a Paragraph Ranker mechanism, which uses bidirectional LSTMs to compute representations of passages and questions and measures similarities between the passages and questions by dot product to score each passage. 
\par For joint training, Reinforced Ranker-Reader ($ R^{3} $), proposed by Wang et al. \cite{wang2018r}, is the representative model. In $R^{3}$, match-LSTM \cite{wang2016machine} is applied to compute the similarity between the question and each passage to obtain document representations, which are later fed to the ranker and the reader. In the ranker module, reinforcement learning is used to select the most relevant passage, while the function of the reader is to predict the answer span from this selected passage. These two tasks are trained jointly to mitigate error propagation caused by wrong document retrieval.
\par However, retrieval components in the above models have low efficiency. For example, DrQA~\cite{chen2017reading} simply uses traditional IR approaches in the retrieval component, and $R^{3}$~\cite{wang2018r} applies question-dependent passage representations to rank the passages. The computational complexity increases as the document corpus becomes larger. To accelerate the retrieval process, Das~et~al.~\cite{das2019multi} propose a fast and efficient retrieval method that represents passages independent from questions and stores outputs offline. When given the question, the model computes a fast inner product to measure similarities between passages and questions. Then the top-ranked passages are fed to the reader to extract answers. Another unique characteristic of this work is the iterative interaction between the retriever and the reader. They introduce a gated recurrent unit to reformulate query representations to account for the state of the reader and the original query. The new query representations are then used to retrieve other relevant passages, which facilitates the reread process across the corpus.
\par In the multi-passage setting, there may be more than one possible answer, among which some are not correct answers to the question. Instead of selecting the first match span as the correct answer, Pang et al. \cite{pang2019has} propose three heuristic methods. The RAND operation treats all answer spans equally and chooses one randomly, while the MAX operation chooses the one with maximum probability and can be used if there are noisy paragraphs. The SUM operation assumes that more than one span can be regarded as ground-truth and sums all span probabilities together. Similar to the MAX operation, Clark~and Gardner \cite{clark2018simple} regard all labeled answer spans as correct and, inspired by attention sum reader \cite{kadlec2016text}, they use a summed objective function to choose the one with maximum probability to be the correct answer. In contrast, Lin et al. \cite{lin2018denoising} introduce a fast paragraph selector to filter out passages with wrong answer labels before feeding them to the reader module. They use multi-layer perceptron or RNNs to obtain hidden representations of passages and questions, respectively. \mbox{In addition}, a self-attention operation is applied to the questions to illustrate their different importance. Then the similarity between passages and questions is calculated and the most similar ones are chosen as relevant passages to be fed to the reader module.
\par Wang et al. \cite{Wang2018evidence} see the significance of evidence aggregation in multi-passage MRC tasks. \mbox{In their} point of view, on the one hand, correct answers have more evidence across different passages. \mbox{On the other} hand, some questions require various aspects of evidence to answer. To make full use of multiple pieces of evidence, they propose strength-based and coverage-based re-rankers. In the first mechanism, the answer with the most occurrences among the candidates is chosen to be the correct one. The second re-ranker concatenates all passages that contain candidate answers as a new context and feeds that to the reader to obtain the answer which aggregates different aspects of evidence.
\par To sum up, compared to MRC tasks, multi-passage MRC is much closer to real-world application. With several documents given as resources, there is more evidence for answer prediction, thus even if the question is complicated, the model can give the answer fairly well. Related document retrieval is important in multi-passage MRC, and evidence aggregated from documents may be complementary or contradictory. Hence, free answering, in which answers are not limited to the subsequence in the original context, is common in multi-passage MRC tasks. Taking advantage of multiple documents and generating answers with the proper logic and clear semantics to answer questions still has a long way to go. 
\subsection{Conversational Machine Reading Comprehension}
\par MRC requires answering a question based on the understanding of a given passage, with questions usually isolated from each other. However, the most natural way that people acquire knowledge is via a series of interrelated question-and-answer processes. When given a document, someone will ask a question, and someone else gives an answer. Then based on the answer, a related question is asked for deeper understanding. This process is performed iteratively, which can be regarded as a multi-turn conversation. After incorporating conversation into MRC, conversational machine reading comprehension (CMRC) has become the research hot spot.
\par In contrast to MRC, conversation history $H$ also acts as part of the context to help with answer prediction in CMRC. The definition of CMRC can be formulated as shown in Table \ref{d_conversational}:
\vspace{12pt}
\begin{table}[h]
	\renewcommand{\arraystretch}{1.5}
	\centering
	\caption{Definition of conversational machine reading comprehension.}
	\label{d_conversational}
	\begin{tabular}{p{12cm}}
		\noalign{\hrule height 1pt}
		\bfseries{Conversational Machine Reading Comprehension}\\
		\hline
		\rowcolor{tabcolor}Given the context $C$, the conversation history with previous questions and answers $H=\{Q_{1},A_{1},\cdots,Q_{i-1},A_{i-1}\}$ and the current question $Q_{i}$, the CMRC task is to predict the correct answer $A_{i} $ by learning the function $\mathcal{F}$ such that $A_{i}=\mathcal{F}(C,H,Q_{i})$.\\
		\noalign{\hrule height 1pt}
	\end{tabular}
\end{table}
\vspace{12pt}

\par Many researchers have tried to create new datasets with given passages and a series of conversations to satisfy the requirements of CMRC tasks. Reddy et al. \cite{reddy2019coqa} release CoQA, a CMRC dataset with 8000 conversations about passages in seven domains. In CoQA, a questioner asks questions based on a given passage and an answerer gives answers, simulating a conversation between two humans when reading the passage. There is no limit to the answer forms in CoQA, which requires more contextual reasoning. Similarly, Choi et al. \cite{choi2018quac} introduce QuAC for question answering in context. Compared to CoQA, in QuAC, the passage is given only to the answerer, and the questioner asks the questions based on the title of the passages. The answerer answers the questions with subsequences of the original passage and determines whether the questioner can ask a follow-up question. Ma et al. \cite{ma2018challenging} extend cloze test tasks to the conversational setting. They use dialogues between characters selected from transcripts of the TV show \textit{Friends} to generate related context and ask to fill in the blanks with character names according to utterances and context. In contrast to the above two datasets, it is aimed at multi-party dialogue and pays attention to the doers of some actions. \mbox{To illustrate} the CMRC task more specifically, some examples in CoQA datasets are presented \mbox{in Table \ref{CQA}.}

\begin{table}[h]
	\renewcommand{\arraystretch}{1.2}
	\centering
	\caption{\upshape An example of the CoQA dataset}
	\label{CQA}
	\begin{tabular}{ll}
		\noalign{\hrule height 1pt}
		\rowcolor{tabcolor}\multicolumn{2}{l}{\bfseries{CMRC}}\\
		\hline
		Passage: & \color{red}Jessica \color{black} went to sit in her rocking chair. Today was \color{red}her \color{black} birthday\\ 
		& and she was turning 80. Her granddaughter Annie was coming\\
		&  over in the afternoon and Jessica was very excited to see her.\\
		& Her daughter Melanie and Melanie’s husband Josh were coming\\
		& as well.\\
		\hline
		Question 1: & Who had a birthday?\\
		Answer 1 : & \color{red}Jessica\color{black} \\
		\hline
		Question 2: & How old would \color{red}she \color{black} be?\\
		Answer 2 : & 80 \\ 
		\hline
		Question 3: & Did she plan to have any \color{blue}visitors\color{black}?\\
		Answer 3 : & Yes \\
		\hline
		Question 4: & \color{blue}How many?\color{black}\\
		Answer 4 : & Three \\
		\hline
		Question 5: & \color{blue}Who?\color{black}\\
		Answer 5: & Annie, Melanie and Josh \\
		\noalign{\hrule height 1pt}
	\end{tabular}
\end{table}

\par CMRC brings about some new challenges compared to MRC:
\begin{itemize}[itemindent=2em,listparindent=2em,leftmargin=0pt,itemsep=10pt,parsep=10pt]
	\item[-] Conversational History
	\par In MRC tasks, questions and answers are based on given passages and questions are independent from the previous question-answering process. In contrast to that, conversational history plays an  important role in CMRC. A follow-up question may be closely related to prior questions and answers. More specifically, as shown in Table \ref{CQA}, question 4 and 5 are relevant to question 3. Moreover, answer 3 can be a verification for answer 5. To face this challenge, dialogue pairs as conversational history are also fed to the CMRC systems as inputs.
	\item[-] Coreference Resolution
	\par Coreference resolution is a traditional task in natural language processing, and it is even more challenging in CMRC. The coreference phenomenon not only may occur in the context, but may appear in the question-and-answer sentences as well. Coreference can be sorted into two kinds, explicit and implicit. With explicit coreference, there are explicit markers, such as some personal pronouns. \mbox{For instance}, to answer question 1 \textit{Who had a birthday} in Table \ref{CQA}, the model has to figure out that \textit{``her''} in \textit{Today was her birthday} refers to \textit{Jessica}. Similarly, the understanding of question 2 is based on the knowledge that \textit{she} means \textit{Jessica}. By comparison, implicit coreference without explicit markers is much harder to figure out. Short questions with certain intentions that implicitly refer to previous content are a kind of implicit coreference. For example, to figure out the complete expression of question 4 (\textit{How many are the visitors?}), the model should extract the correlation between \mbox{question 4 and 3.}
\end{itemize}
\vspace{-10pt}
\par Over the past two years, some researchers have made an effort to tackle the above new challenges in CMRC tasks. Reddy et al. \cite{reddy2019coqa} propose a hybrid model, DrQA+PGNet, which combines the sequence-to-sequence and machine reading comprehension models to extract and generate answers. To integrate information on conversational history, they treat previous question-answer pairs as a sequence and append them to the context. Yatskar et al. \cite{yatskar2019qualitative} use an improved MRC model, Bi-DAF++ with ELMo \cite{peters2018deep} to answer questions based on the given context and conversational history. Rather than encoding previous dialogue information on the context representation, they label answers to previous questions in the context. Instead of simply concatenating previous question-answer pairs as inputs, Huang et al. \cite{huang2018flowqa} introduce a flow mechanism to deeply understand conversational history, which encodes hidden context representations during the process of answering previous questions. Similar to Reddy et al. \cite{reddy2019coqa}, Zhu et al. \cite{zhu2018sdnet} append previous question-answer pairs to current questions, but in order to find out related conversational history, they employ additional self-attention on questions.
\par CMRC tasks that incorporate conversation into MRC are in line with the general process that humans understand one thing in the real world. Although researchers have been aware of the significance of conversational information and succeeded in representing conversational history, there is little work on coreference resolution. If the coreference cannot be figured out correctly, it will result in performance degradation. The common coreference phenomena in CMRC make this task far more~challenging.
\section{Open Issues} \label{s6}
Based on the analysis of the literature cited in this article, we observe that there are some open issues that remain unsolved in neural MRC, some of which may have been also discussed in related research, such as machine inference and open-domain QA. The most important issue in neural MRC is that the machine does not truly understand the given text, as existing models mainly rely on semantic matching to answer question. Experiments performed by Kaushik and Lipton \cite{kaushik2018much} show that some MRC models perform unexpectedly well when being provided with just passages or questions. Although researchers have already made great strides in neural MRC, there is still a giant gap between MRC and human comprehension in the following aspects:

\begin{itemize}[itemindent=2em,listparindent=2em,leftmargin=0pt,itemsep=10pt,parsep=10pt]
	\item[-] Limitation of Given Context
	\par  Similar to reading comprehension on language proficiency tests, machine reading comprehension asks the machine to answer questions based on the given context. Such context is a necessity in MRC tasks, but restricts the application. In the real world, machines are not expected to help students with their reading comprehension exams but make question-answering or dialogue systems smarter. Efforts made in multi-passage MRC research have somewhat broken the limitation of given context, but there is still a long way to go as how to find the most relevant resources for MRC systems effectively determines the performance of answer prediction. It calls for a deeper combination of information retrieval and machine reading comprehension in the future.
	\item[-] Robustness of MRC Systems
	\par As Jia and Liang \cite{jia2017adversarial} point out, most existing MRC models based on word overlap are weak to adversarial question-answer pairs. For SQuAD, they add distracting sentences to the given context, which have semantic overlap with the question and might cause confusion but do not contradict the correct answer. With such adversarial examples, the performance of MRC systems drops dramatically, which reflects that machines cannot really understand natural language. Although the introduction of answer verification components can alleviate the side-effect of plausible answers in a way, the~robustness of MRC systems should be enhanced to face the challenges of adversarial circumstances. 
	\item[-] Incorporation of External Knowledge
	\par As an essential component of human intelligence, accumulated common sense or background knowledge is usually used in human reading comprehension. To mimic this, a knowledge-based MRC is proposed to improve the performance of machine reading with external knowledge. However, how to effectively introduce external knowledge and make full use of it remains an ongoing challenge. On the one hand, the structure of knowledge stored in knowledge bases is so different from the text in context and questions that it is difficult to integrate them. On the other hand, the performance of knowledge-based MRC is closely related to the quality of the knowledge base. Constructing of a knowledge base is time-consuming, requiring considerable human efforts. In addition, knowledge in knowledge bases is sparse, and most of the time, relevant external knowledge cannot be found directly to support answer prediction and further reasoning is required. The effective fusion of knowledge graphs and machine reading comprehension needs to be further investigated.
	\item[-] Lack of Inference Ability
	\par As mentioned before, most existing MRC systems are mainly based on semantic matching between the context and the question to give the answer, which results in MRC models being incapable of reasoning. As an example given by Liu et al. \cite{liu2019r} shows, given the context that \textit{five people on board and two people on the ground died}, the machine cannot infer the correct answer \textit{seven} to the question \textit{how many people died} because of the lack of inference ability. How to enable the machine with inference ability still requires further research.
	\item[-] Difficulty in Interpretation
	\par Although existing MRC systems excel in a variety of tasks, they still work in a black box manner, i.e., no information is provided about how exactly they predict the answer. The lack of interpretability is a major drawback in applications such as healthcare, where the rationale for a model's output is a requirement for trust. However, the release of some MRC datasets such as HotpotQA \cite{yang2018hotpotqa} may contribute to the interpretability of MRC systems. HotpotQA, consisting of questions requiring inference over multiple supporting documents, labels sentences useful for reasoning so that the MRC systems are allowed to explain the predictions with such information. In addition, with development in explainable artificial intelligence (XAI) \cite{shrikumar2016not,lipton2016mythos,montavon2017explaining}, the MRC research field has good potential for improving the trust and transparency of MRC systems and in turn making applications using MRC techniques more practical.
\end{itemize}
\section{Conclusions}
This article has presented a comprehensive survey on the progress of neural machine reading comprehension. Based on a thorough analysis of recent work, we give the specific definitions of MRC tasks and compare them in depth. The general architecture of neural MRC models is decomposed into four modules, and prominent approaches used in each module are introduced in detail. Representative datasets as well as evaluation metrics are described according to different MRC tasks. In addition, considering the limitations of MRC, we shed light on some new trends and discuss open issues in this research field.

\appendix
\section{Glossary of Items Mentioned} \label{appendix}

\begin{longtable}[c]{ll}
	\label{glossary}\\
	\toprule
	\textbf{Term} & \textbf{Definition}\\
	\midrule
	\textbf{Curse of Dimensionality} & \makecell[l]{A phenomenon that appears with high-dimensional\\ data, where the computational complexity grows\\ exponentially with the number of dimensions increasing.}\\
	\midrule
	\textbf{Transfer Learning} & \makecell[l]{A method of using knowledge from a related task\\ that has already been learned to learn new tasks.}\\
	\midrule
	\textbf{Coattention}& \makecell[l]{A kind of bidirectional attention mechanism designed\\ by Xiong et al.  \cite{xiong2016dynamic} which can attend to the context\\ and question simultaneously.}\\
	\midrule
	\textbf{Dynamic Decoder}& \makecell[l]{A mechanism proposed by Xiong et al. \cite{xiong2016dynamic} that is\\ based on LSTMs, dynamically predicting the start\\ and end positions of the answer.}\\
	\midrule
	\textbf{Zero-Shot Setting} & \makecell[l]{A setting where labeled examples for training are not\\ enough.}\\
	\midrule
	\textbf{Masked Language Model} & \makecell[l]{A model designed by Devlin et al. \cite{devlin2019bert}, which randomly\\ mask some tokens from the input, and requires predicting\\ the masked word based on its context.}\\
	\midrule
	\textbf{Next-Sentence Prediction} & \makecell[l]{A task designed by Devlin et al. \cite{devlin2019bert}, judging whether\\ sentence $B$ follows sentence $A$.}\\
	\midrule
	\textbf{Out-of-Vocabulary}&\makecell[l]{Unknown words that appear in the training examples\\ but not in the pre-defined vocabulary.} \\
	\midrule
	\textbf{Fine-Grained Gating} & \makecell[l]{A mechanism proposed by Yang et al. \cite{yang2016words} which uses\\ linguistic features, such as POS tags and name-entity\\ tags, as a gate to control the amount of information\\ flowing to word-level and character-level representations.}\\
	\midrule
	\textbf{Gradient Explosion} & \makecell[l]{A problem where a considerable gradient accumulate\\ as it moves backward through the layers while training\\ a deep neural network.}\\
	\midrule
	\textbf{Gradient Vanishing} & \makecell[l]{A problem where the gradient tends to get smaller as\\ it moves backward through the layers while training a \\deep neural network.}\\
	\midrule
	\textbf{Multi-Head Attention} & \makecell[l]{A mechanism proposed by Vaswani et al. \cite{vaswani2017attention} using\\ different linear projections to project queries, keys\\ and values more than once when performing attention\\ function.}\\
	\midrule
	\textbf{Self-Attention} & \makecell[l]{Also \textit{Intra-Attention}, an attention mechanism compu-\\ting attention weights in a single sequence to highlight\\ importance of different positions in the sequence.}\\
	\midrule
	\textbf{Residual Connections} & \makecell[l]{A mechanism that skip some layer connections to facili-\\tate single propagation and mitigate gradient degradation.}\\
	\midrule
	\textbf{Back-Propagation} & \makecell[l]{A shortened form of \textit{backward propagation of errors}, a\\ method of training neural networks with errors propaga-\\ting backwards.}\\
	\midrule
	\textbf{Textual Entailment} & \makecell[l]{A task to judge the relationship between text fragments\\ whether a truth of one text fragment follows from another.}\\
	\midrule
	\textbf{Pointer Networks} & \makecell[l]{A neural architecture proposed by Vinyals et al. \cite{vinyals2015pointer} which\\ uses attention mechanism as a pointer to select several\\ discrete tokens in input sequence as the output.}\\
	\midrule
	\textbf{Maxout Networks} & \makecell[l]{A model proposed by Goodfellow et al. \cite{goodfellow2013maxout} whose output\\ is the maximum of a set of inputs, to improve optimization\\ and model averaging with dropout.} \\
	\midrule
	\textbf{Highway Networks} & \makecell[l]{An architecture proposed by Srivastava et al. \cite{srivastava2015highway} which\\ contains information highways to accelerate the training\\ process of gradient-based deep neural networks.}\\
	\midrule
	\textbf{Multi-Task Learning} & \makecell[l]{A kind of transfer learning which solves multiple tasks\\ at the same time.}\\
	\midrule
	\textbf{Key-Value Memory Networks} & \makecell[l]{A neural network proposed by Miller et al. \cite{miller2016key} which\\ stores memory as key-value pairs for easy access.}\\
	\midrule
	\textbf{Data Enrichment} & \makecell[l]{The process of enriching the amount of raw data.}\\
	\bottomrule
\end{longtable}

\bibliographystyle{plain}
\bibliography{mrc_survey}

\end{document}